\documentclass[10pt,twocolumn,letterpaper]{article}
\pdfoutput=1
\newlength\minalignvsep

\usepackage{cvpr}
\usepackage{times}
\usepackage{epsfig}
\usepackage{graphicx}
\usepackage{amsmath,amssymb,amsthm}
\usepackage{pbox}
\usepackage{epstopdf}
\usepackage{subfigure}
\usepackage{xspace}
\usepackage{comment}
\usepackage{lipsum}
\usepackage{enumitem}

\newcommand{\cmt}[2]{[#1: #2]}
\newcommand{\todo}[1]{\cmt{{\bf TODO}}{{\bf \color{red} #1}}}

\newcommand{\mypara}{\vspace*{-15pt}\paragraph}

\newcommand{\denselist}{\itemsep 0pt\parsep=0pt\partopsep 0pt\vspace{-2pt}}
\newcommand{\bitem}{\begin{itemize}\denselist}
\newcommand{\eitem}{\end{itemize}}
\newcommand{\benum}{\begin{enumerate}\denselist}
\newcommand{\eenum}{\end{enumerate}}
\newcommand{\bdescr}{\begin{description}\denselist}
\newcommand{\edescr}{\end{description}}

\newtheorem{theorem}{Theorem}

\newcommand{\myvec}[1]{\mathbf #1}
\newcommand\blfootnote[1]{%
  \begingroup
  \renewcommand\thefootnote{}\footnote{#1}%
  \addtocounter{footnote}{-1}%
  \endgroup
}
\newtheorem{innercustomthm}{Theorem}
\newenvironment{customthm}[1]
  {\renewcommand\theinnercustomthm{#1}\innercustomthm}
  {\endinnercustomthm}
  
\usepackage[pagebackref=true,breaklinks=true,letterpaper=true,colorlinks,bookmarks=false]{hyperref}

\cvprfinalcopy 

\def\cvprPaperID{201} 

\ifcvprfinal\fi
\begin{document}

\title{PointNet: Deep Learning on Point Sets for 3D Classification and Segmentation}


\author{Charles R. Qi*\qquad Hao Su* \qquad Kaichun Mo \qquad Leonidas J. Guibas\\Stanford University}

\maketitle

\begin{abstract}

Point cloud is an important type of geometric data structure. Due to its irregular format, most researchers transform such data to regular 3D voxel grids or collections of images. This, however, renders data unnecessarily voluminous and causes issues. In this paper, we design a novel type of neural network that directly consumes point clouds, which well respects the permutation invariance of points in the input.  Our network, named PointNet, provides a unified architecture for applications ranging from object classification, part segmentation, to scene semantic parsing. Though simple, PointNet is highly efficient and effective. Empirically, it shows strong performance on par or even better than state of the art. Theoretically, we provide analysis towards understanding of what the network has learnt and why the network is robust with respect to input perturbation and corruption.


\end{abstract}

\section{Introduction}
\blfootnote{* indicates equal contributions.}
\label{sec:intro}


In this paper we explore deep learning architectures capable of reasoning about 3D geometric data such as point clouds or meshes. Typical convolutional architectures require highly regular input data formats, like those of image grids or 3D voxels, in order to perform weight sharing and other kernel optimizations. Since point clouds or meshes are not in a regular format, most researchers typically transform such data to regular 3D voxel grids or collections of images (e.g, views) before feeding them to a deep net architecture. This data representation transformation, however, renders the resulting data unnecessarily voluminous --- while also introducing quantization artifacts that can obscure natural invariances of the data. 

For this reason we focus on a different input representation for 3D geometry using simply point clouds -- and name our resulting deep nets {\em PointNets}. Point clouds are simple and unified structures that avoid the combinatorial irregularities and complexities of meshes, and thus are easier to learn from. The PointNet, however, still has to respect the fact that a point cloud is just a set of points and therefore invariant to permutations of its members, necessitating certain symmetrizations in the net computation. Further invariances to rigid motions also need to be considered.

\begin{figure}
    \centering
    \includegraphics[width=\linewidth]{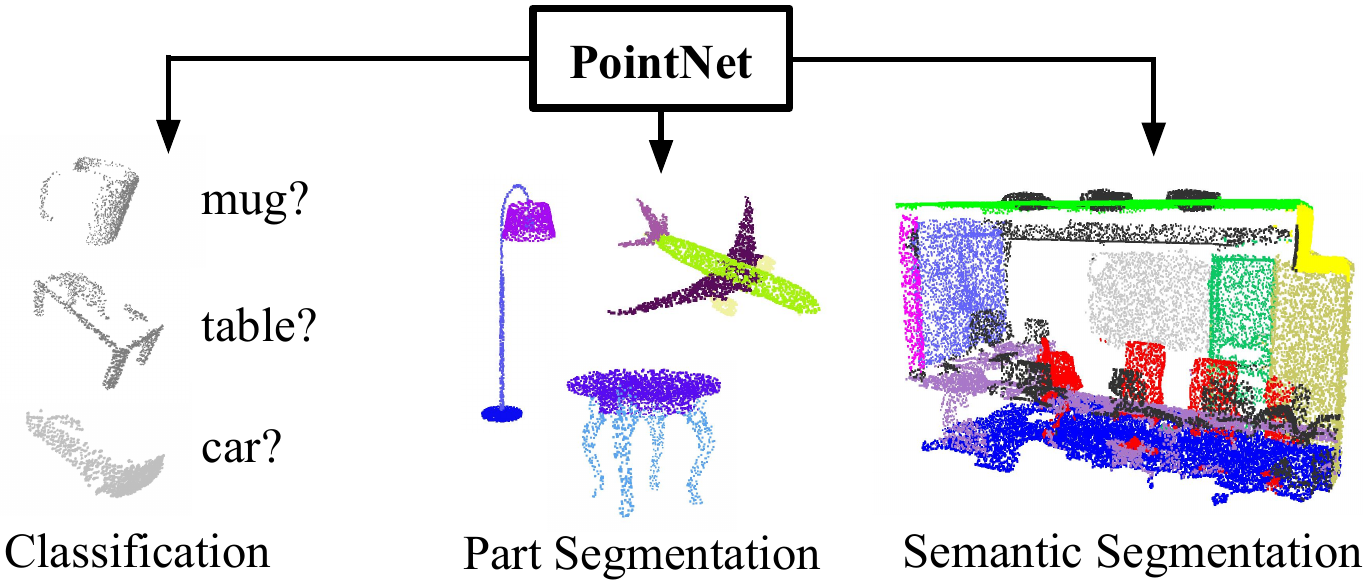}
    \caption{\textbf{Applications of PointNet.} We propose a novel deep net architecture that consumes raw point cloud (set of points) without voxelization or rendering. It is a unified architecture that learns both global and local point features, providing a simple, efficient and effective approach for a number of 3D recognition tasks.}
    \label{fig:teaser}
\end{figure}

Our PointNet is a unified architecture that directly takes point clouds as input and outputs either class labels for the entire input or per point segment/part labels for each point of the input. 
The basic architecture of our network is surprisingly simple as in the initial stages each point is processed identically and independently. In the basic setting each point is represented by just its three coordinates $(x, y, z)$. Additional dimensions may be added by computing normals and other local or global features. 

Key to our approach is the use of a single symmetric function, max pooling. Effectively the network learns a set of optimization functions/criteria that select interesting or informative points of the point cloud and encode the reason for their selection. The final fully connected layers of the network aggregate these learnt optimal values into the global descriptor for the entire shape as mentioned above (shape classification) or are used to predict per point labels (shape segmentation). 

Our input format is easy to apply rigid or affine transformations to, as each point transforms independently. Thus we can add a data-dependent spatial transformer network that attempts to canonicalize the data before the PointNet processes them, so as to further improve the results. 

We provide both a theoretical analysis and an experimental evaluation of our approach. We show that our network can approximate any set function that is continuous. More interestingly, it turns out that our network learns to summarize an input point cloud by a sparse set of key points, which roughly corresponds to the skeleton of objects according to visualization. The theoretical analysis provides an understanding why our PointNet is highly robust to small perturbation of input points as well as to corruption through point insertion (outliers) or deletion (missing data).

On a number of benchmark datasets ranging from shape classification, part segmentation to scene segmentation, we experimentally compare our PointNet with state-of-the-art approaches based upon multi-view and volumetric representations. Under a unified architecture, not only is our PointNet much faster in speed, but it also exhibits strong performance on par or even better than state of the art.

The key contributions of our work are as follows:
\bitem
\item We design a novel deep net architecture suitable for consuming unordered point sets in 3D;
\item We show how such a net can be trained to perform 3D shape classification, shape part segmentation and scene semantic parsing tasks;
\item We provide thorough empirical and theoretical analysis on the stability and efficiency of our method;
\item We illustrate the 3D features computed by the selected neurons in the net and develop intuitive explanations for its performance.
\eitem

The problem of processing unordered sets by neural nets is a very general and fundamental problem -- we expect that our ideas can be transferred to other domains as well.

\begin{figure*}[th!]
    \centering
    \includegraphics[width=0.9\linewidth]{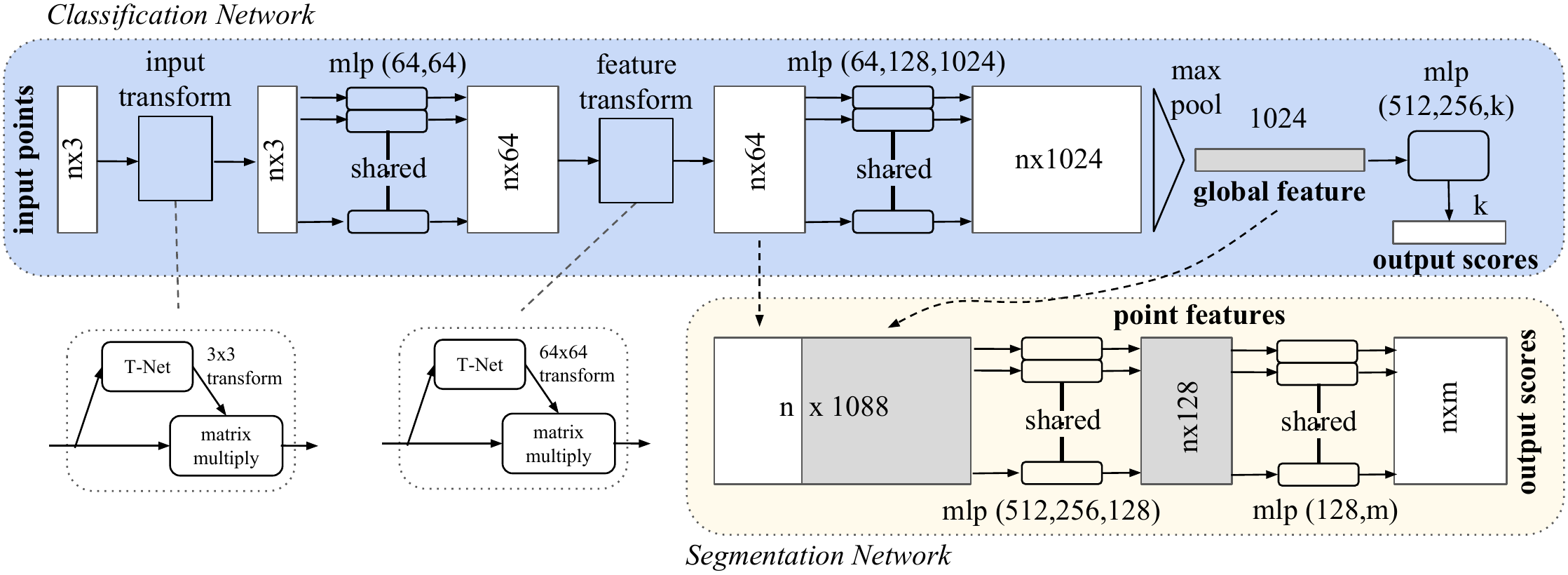}
    \caption{\textbf{PointNet Architecture.} The classification network takes $n$ points as input, applies input and feature transformations, and then aggregates point features by max pooling. The output is classification scores for $k$ classes. The segmentation network is an extension to the classification net. It concatenates global and local features and outputs per point scores. ``mlp'' stands for multi-layer perceptron, numbers in bracket are layer sizes. Batchnorm is used for all layers with ReLU. Dropout layers are used for the last mlp in classification net.}
    \label{fig:pointnet_arch}
\end{figure*}

\section{Related Work}
\label{sec:related}

\paragraph{Point Cloud Features}
Most existing features for point cloud are handcrafted towards specific tasks. Point features often encode certain statistical properties of points and are designed to be invariant to certain transformations, which are typically classified as intrinsic~\cite{aubry2011wave, sun2009concise, bronstein2010scale} or extrinsic ~\cite{rusu2008aligning, rusu2009fast, ling2007shape, johnson1999using, chen2003visual}.  They can also be categorized as local features and global features. For a specific task, it is not trivial to find the optimal feature combination.

\paragraph{Deep Learning on 3D Data}

3D data has multiple popular representations, leading to various approaches for learning. 
\emph{Volumetric CNNs:}~\cite{wu20153d, maturana2015voxnet, qi2016volumetric} are the pioneers applying 3D convolutional neural networks on voxelized shapes. However, volumetric representation is constrained by its resolution due to data sparsity and computation cost of 3D convolution. FPNN~\cite{li2016fpnn} and Vote3D~\cite{wang2015voting} proposed special methods to deal with the sparsity problem; however, their operations are still on sparse volumes, it's challenging for them to process very large point clouds. 
\emph{Multiview CNNs:}~\cite{su15mvcnn, qi2016volumetric} have tried to render 3D point cloud or shapes into 2D images and then apply 2D conv nets to classify them. With well engineered image CNNs, this line of methods have achieved dominating performance on shape classification and retrieval tasks~\cite{savvashrec}. However, it's nontrivial to extend them to scene understanding or other 3D tasks such as point classification and shape completion. 
\emph{Spectral CNNs:} Some latest works~\cite{bruna2013spectral, masci2015geodesic} use spectral CNNs on meshes. However, these methods are currently constrained on manifold meshes such as organic objects and it's not obvious how to extend them to non-isometric shapes such as furniture. 
\emph{Feature-based DNNs:}~\cite{fang20153d,guo20153d} firstly convert the 3D data into a vector, by extracting traditional shape features and then use a fully connected net to classify the shape. We think they are constrained by the representation power of the features extracted.

\paragraph{Deep Learning on Unordered Sets}

From a data structure point of view, a point cloud is an unordered set of vectors. While most works in deep learning focus on regular input representations like sequences (in speech and language processing), images and volumes (video or 3D data), not much work has been done in deep learning on point sets.

One recent work from Oriol Vinyals et al~\cite{vinyals2015order} looks into this problem. They use a read-process-write network with attention mechanism to consume unordered input sets and show that their network has the ability to sort numbers. However, since their work focuses on generic sets and NLP applications, there lacks the role of geometry in the sets.


\section{Problem Statement}
\label{sec:problem}

We design a deep learning framework that directly consumes unordered point sets as inputs. 
A point cloud is represented as a set of 3D points $\{P_i| \ i=1,...,n\}$, where each point $P_i$ is a vector of its $(x,y,z)$ coordinate plus extra feature channels such as color, normal etc. For simplicity and clarity, unless otherwise noted, we only use the $(x,y,z)$ coordinate as our point's channels. 

For the object classification task, the input point cloud is either directly sampled from a shape or pre-segmented from a scene point cloud. Our proposed deep network 
outputs $k$ scores for all the $k$ candidate classes.
For semantic segmentation, the input can be a single object for part region segmentation, or a sub-volume from a 3D scene for object region segmentation. Our model will output $n \times m$ scores for each of the $n$ points and each of the $m$ semantic sub-categories.

\section{Deep Learning on Point Sets}
The architecture of our network (Sec~\ref{sec:pointnet_arch}) is inspired by the properties of point sets in $\mathbb{R}^n$ (Sec~\ref{sec:point_set_property}). 

\subsection{Properties of Point Sets in $\mathbb{R}^n$}
\label{sec:point_set_property}
Our input is a subset of points from an Euclidean space. It has three main properties:

\bitem
\item Unordered.
Unlike pixel arrays in images or voxel arrays in volumetric grids, point cloud is a set of points without specific order. In other words, a network that consumes $N$ 3D point sets needs to be invariant to $N!$ permutations of the input set in data feeding order. 
\item Interaction among points. The points are from a space with a distance metric. It means that points are not isolated, and neighboring points form a meaningful subset. Therefore, the model needs to be able to capture local structures from nearby points, and the combinatorial interactions among local structures. 
\item Invariance under transformations.
As a geometric object, the learned representation of the point set should be invariant to certain transformations. For example, rotating and translating points all together should not modify the global point cloud category nor the segmentation of the points.
\eitem


\subsection{PointNet Architecture}
\label{sec:pointnet_arch}

Our full network architecture is visualized in Fig~\ref{fig:pointnet_arch}, where the classification network and the segmentation network share a great portion of structures. Please read the caption of Fig~\ref{fig:pointnet_arch} for the pipeline.

Our network has three key modules: the max pooling layer as a symmetric function to aggregate information from all the points, a local and global information combination structure, and two joint alignment networks that align both input points and point features.

We will discuss our reason behind these design choices in separate paragraphs below. 

%
%
%
%

\paragraph{Symmetry Function for Unordered Input}
In order to make a model invariant to input permutation, three strategies exist: 1) sort input into a canonical order; 2) treat the input as a sequence to train an RNN, but augment the training data by all kinds of permutations; 3) use a simple symmetric function to aggregate the information from each point. Here, a symmetric function takes $n$ vectors as input and outputs a new vector that is invariant to the input order. For example, $+$ and $*$ operators are symmetric binary functions. 

While sorting sounds like a simple solution, in high dimensional space there in fact does not exist an ordering that is stable w.r.t. point perturbations in the general sense. This can be easily shown by contradiction. If such an ordering strategy exists, it defines a bijection map between a high-dimensional space and a $1d$ real line. It is not hard to see, to require an ordering to be stable w.r.t point perturbations is equivalent to requiring that this map preserves spatial proximity as the dimension reduces, a task that cannot be achieved in the general case. Therefore, sorting does not fully resolve the ordering issue, and it's hard for a network to learn a consistent mapping from input to output as the ordering issue persists. As shown in experiments (Fig~\ref{fig:order_invariant}), we find that applying a MLP directly on the sorted point set performs poorly, though slightly better than directly processing an unsorted input.

The idea to use RNN considers the point set as a sequential signal and hopes that by training the RNN with randomly permuted sequences, the RNN will become invariant to input order. However in ``OrderMatters''~\cite{vinyals2015order} the authors have shown that order does matter and cannot be totally omitted. While RNN has relatively good robustness to input ordering for sequences with small length (dozens), it's hard to scale to thousands of input elements, which is the common size for point sets. Empirically, we have also shown that model based on RNN does not perform as well as our proposed method (Fig~\ref{fig:order_invariant}).

Our idea is to approximate a general function defined on a point set by applying a symmetric function on transformed elements in the set: 
\begin{align}
    f(\{x_1, \dots, x_n\})\approx g(h(x_1), \dots, h(x_n)),
    \label{eq:approx}
\end{align}
where $f:2^{\mathbb{R}^N} \rightarrow \mathbb{R}$, $h: \mathbb{R}^N\rightarrow \mathbb{R}^K$ and $g:\underbrace{\mathbb{R}^K\times \dots \times \mathbb{R}^K}_n \rightarrow \mathbb{R}$ is a symmetric function.

Empirically, our basic module is very simple: we approximate $h$ by a multi-layer perceptron network and $g$ by a composition of a single variable function and a max pooling function. This is found to work well by experiments. Through a collection of $h$, we can learn a number of $f$'s to capture different properties of the set. 

While our key module seems simple, it has interesting properties (see Sec~\ref{sec:visualizing_pointnet}) and can achieve strong performace (see Sec~\ref{sec:application}) in a few different applications. Due to the simplicity of our module, we are also able to provide theoretical analysis as in Sec~\ref{sec:theory}.

\paragraph{Local and Global Information Aggregation}
The output from the above section forms a vector $[f_1, \dots, f_K]$, which is a global signature of the input set. We can easily train a SVM or multi-layer perceptron classifier on the shape global features for classification. However, point segmentation requires a combination of local and global knowledge. We can achieve this by a simple yet highly effective manner. 

Our solution can be seen in Fig~\ref{fig:pointnet_arch} (\textit{Segmentation Network}). After computing the global point cloud feature vector, we feed it back to per point features by concatenating the global feature with each of the point features. Then we extract new per point features based on the combined point features - this time the per point feature is aware of both the local and global information. 

With this modification our network is able to predict per point quantities that rely on both local geometry and global semantics. For example we can accurately predict per-point normals (fig in supplementary), validating that the network is able to summarize information from the point's local neighborhood. In experiment session, we also show that our model can achieve state-of-the-art performance on shape part segmentation and scene segmentation.
    
\paragraph{Joint Alignment Network}
The semantic labeling of a point cloud has to be invariant if the point cloud undergoes certain geometric transformations, such as rigid transformation. We therefore expect that the learnt representation by our point set is invariant to these transformations. 

A natural solution is to align all input set to a canonical space before feature extraction. Jaderberg et al.~\cite{jaderberg2015spatial} introduces the idea of spatial transformer to align 2D images through sampling and interpolation, achieved by a specifically tailored layer implemented on GPU.

Our input form of point clouds allows us to achieve this goal in a much simpler way compared with~\cite{jaderberg2015spatial}. We do not need to invent any new layers and no alias is introduced as in the image case. We predict an affine transformation matrix by a mini-network (T-net in Fig~\ref{fig:pointnet_arch}) and directly apply this transformation to the coordinates of input points. The mini-network itself resembles the big network and is composed by basic modules of point independent feature extraction, max pooling and fully connected layers. More details about the T-net are in the supplementary.

This idea can be further extended to the alignment of feature space, as well. We can insert another alignment network on point features and predict a feature transformation matrix to align features from different input point clouds. However, transformation matrix in the feature space has much higher dimension than the spatial transform matrix, which greatly increases the difficulty of optimization. We therefore add a regularization term to our softmax training loss. We constrain the feature transformation matrix to be close to orthogonal matrix:
\begin{equation}
    L_{reg} = \|I - AA^T\|_F^2,
\end{equation}
where $A$ is the feature alignment matrix predicted by a mini-network. An orthogonal transformation will not lose information in the input, thus is desired. We find that by adding the regularization term, the optimization becomes more stable and our model achieves better performance.

\subsection{Theoretical Analysis}
\label{sec:theory}
 
\paragraph{Universal approximation} We first show the universal approximation ability of our neural network to continuous set functions. By the continuity of set functions, intuitively, a small perturbation to the input point set should not greatly change the function values, such as classification or segmentation scores.

Formally, let $\mathcal{X}=\{S: S\subseteq [0,1]^m \text{ and } |S|=n\}$,   $f:\mathcal{X}\rightarrow \mathbb{R}$ is a continuous set function on $\mathcal{X}$ w.r.t to Hausdorff distance $d_H(\cdot, \cdot)$, i.e., $\forall \epsilon > 0, \exists \delta >0$, for any $S, S'\in\mathcal{X}$, if $d_H(S, S') < \delta$, then $|f(S)-f(S')|< \epsilon$. Our theorem says that $f$ can be arbitrarily approximated by our network given enough neurons at the max pooling layer, i.e., $K$ in \eqref{eq:approx} is sufficiently large. 

\begin{theorem}
Suppose $f:\mathcal{X}\rightarrow \mathbb{R}$ is a continuous set function w.r.t Hausdorff distance $d_H(\cdot, \cdot)$. $\forall \epsilon > 0$, $\exists$ a continuous function $h$ and a symmetric function $g(x_1, \dots, x_n)=\gamma \circ \mbox{MAX}$, such that for any $S\in\mathcal{X}$,
\begin{align*}
	\left|f(S) - \gamma\left(\underset{x_i\in S}{\mbox{MAX}}\left\{h(x_i)\right\}\right)\right| < \epsilon
\end{align*}
where $x_1, \ldots, x_n$ is the full list of elements in $S$ ordered arbitrarily, $\gamma$ is a continuous function, and $\mbox{MAX}$ is a vector max operator that takes $n$ vectors as input and returns a new vector of the element-wise maximum. 
\end{theorem}
The proof to this theorem can be found in our supplementary material. The key idea is that in the worst case the network can learn to convert a point cloud into a volumetric representation, by partitioning the space into equal-sized voxels. In practice, however, the network learns a much smarter strategy to probe the space, as we shall see in point function visualizations.
\begin{figure}[t!]
    \centering
    \includegraphics[width=0.8\linewidth]{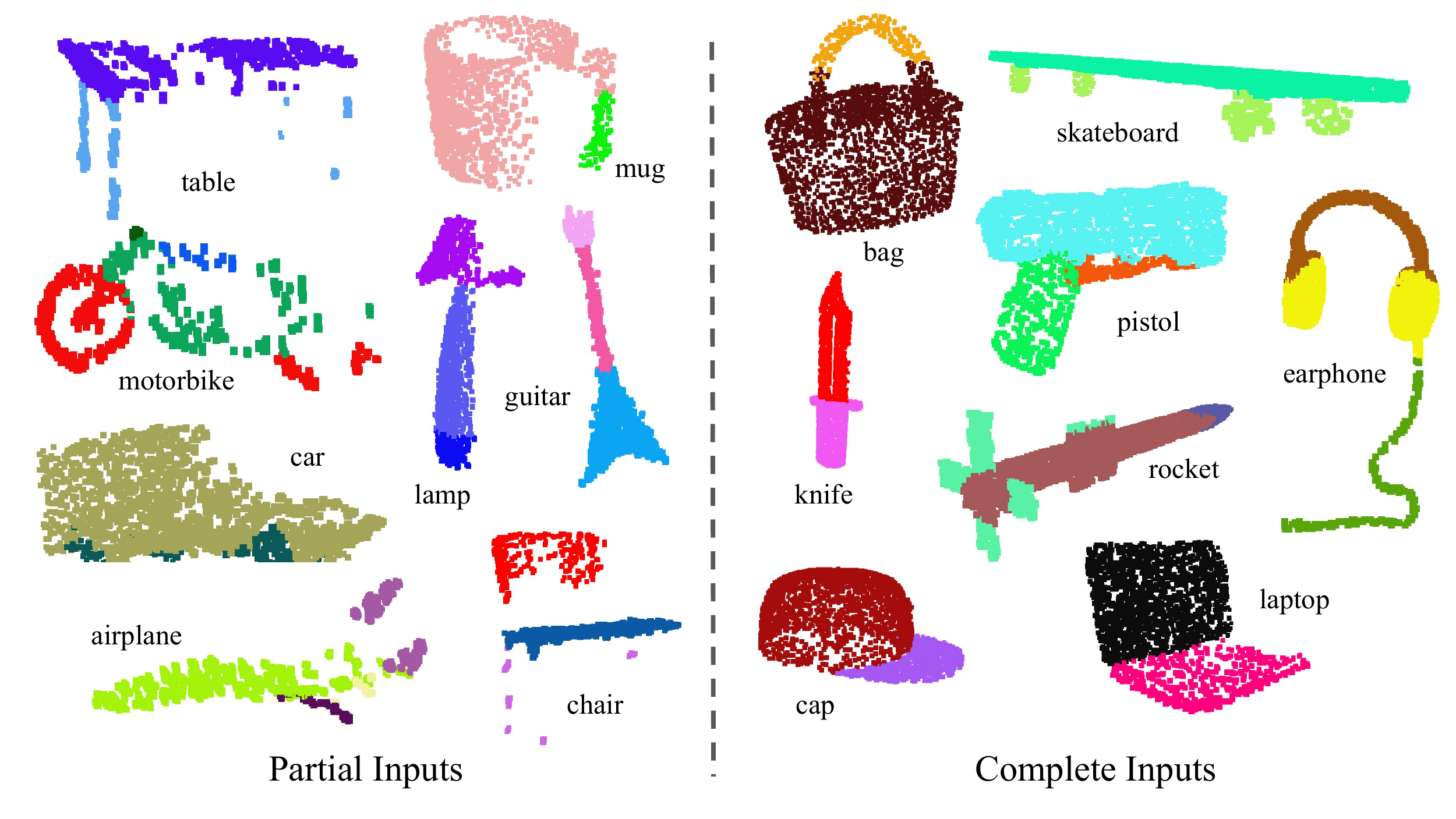}
    \caption{\textbf{Qualitative results for part segmentation.} We visualize the CAD part segmentation results across all 16 object categories. We show both results for partial simulated Kinect scans (left block) and complete ShapeNet CAD models (right block).}
    \label{fig:qualitative_part_segmentation}
\end{figure}

\paragraph{Bottleneck dimension and stability} Theoretically and experimentally we find that the expressiveness of our network is strongly affected by the dimension of the max pooling layer, i.e., $K$ in \eqref{eq:approx}. Here we provide an analysis, which also reveals properties related to the stability of our model. 

We define $\myvec u=\underset{x_i\in S}{\mbox{MAX}}\{h(x_i)\}$ to be the sub-network of $f$ which maps a point set in $[0,1]^m$ to a $K$-dimensional vector. The following theorem tells us that small corruptions or extra noise points in the input set are not likely to change the output of our network:
\begin{theorem}
Suppose $\myvec u:\mathcal{X}\rightarrow \mathbb{R}^K$ such that $\myvec u=\underset{x_i\in S}{\mbox{MAX}}\{h(x_i)\}$ and $f=\gamma \circ \myvec u$. Then, 
\begin{enumerate}[label=(\alph*)]   
    \item $\forall S, \exists~\mathcal{C}_S, \mathcal{N}_S\subseteq \mathcal{X}$,  $f(T)=f(S)$ if  $\mathcal{C}_S\subseteq T\subseteq \mathcal{N}_S$;
    \item $|\mathcal{C}_S| \le K$
\end{enumerate}
\label{thm:thm2}
\end{theorem}
We explain the implications of the theorem. (a) says that $f(S)$ is unchanged up to the input corruption if all points in $\mathcal{C}_S$ are preserved; it is also unchanged with extra noise points up to $\mathcal{N}_S$. (b) says that $\mathcal{C}_S$ only contains a bounded number of points, determined by $K$ in \eqref{eq:approx}. In other words, $f(S)$ is in fact totally determined by a finite subset $\mathcal{C}_S\subseteq S$ of less or equal to $K$ elements. We therefore call $\mathcal{C}_S$ the \emph{critical point set} of $S$ and $K$ the \emph{bottleneck dimension} of $f$. 

Combined with the continuity of $h$, this explains the robustness of our model w.r.t point perturbation, corruption and extra noise points. The robustness is gained in analogy to the sparsity principle in machine learning models. {\bf  Intuitively, our network learns to summarize a shape by a sparse set of key points.} In experiment section we see that the key points form the skeleton of an object.

\section{Experiment}
\begin{table}[t!]
    \small
    \centering
    \begin{tabular}[width=\linewidth]{l|c|c|c|c}
    \hline
    ~               & input        & \#views    & accuracy & accuracy \\ 
    ~ & & & avg. class & overall \\ \hline
    SPH~\cite{kazhdan2003rotation}             & mesh        & - & 68.2         & -  \\ \hline
    3DShapeNets~\cite{wu20153d}     & volume       & 1        & 77.3  & 84.7 \\
    VoxNet~\cite{maturana2015voxnet}          & volume       & 12        & 83.0 & 85.9 \\
    Subvolume~\cite{qi2016volumetric}    & volume       & 20      & 86.0  & \textbf{89.2} \\ \hline
    LFD~\cite{wu20153d}             & image        & 10        & 75.5 & -\\
    MVCNN~\cite{su15mvcnn}           & image        & 80        & \textbf{90.1} & -\\ \hline
    Ours baseline  & point    & -     & 72.6  & 77.4\\
    Ours PointNet   & point   & 1        & 86.2 & \textbf{89.2} \\ \hline
    \end{tabular}
    \caption{\textbf{Classification results on ModelNet40.} Our net achieves state-of-the-art among deep nets on 3D input.}
    \label{tab:classification}
\end{table}
\label{sec:exp}
Experiments are divided into four parts. First, we show PointNets can be applied to multiple 3D recognition tasks (Sec~\ref{sec:application}). Second, we provide detailed experiments to validate our network design (Sec~\ref{sec:arch_analysis}). At last we visualize what the network learns (Sec~\ref{sec:visualizing_pointnet}) and analyze time and space complexity (Sec~\ref{sec:complexity}).

\begin{table*}[th!]
    \small
    \centering
    \begin{tabular}[width=\linewidth]{l|c|p{0.5cm}p{0.4cm}p{0.4cm}p{0.4cm}p{0.5cm}p{0.6cm}p{0.5cm}p{0.5cm}p{0.5cm}p{0.6cm}p{0.6cm}p{0.3cm}p{0.5cm}p{0.6cm}p{0.6cm}p{0.6cm}}
    \hline
    ~        & mean & aero & bag & cap & car & chair & ear & guitar & knife & lamp & laptop & motor & mug & pistol & rocket & skate & table \\ 
    &   & &  &  &  &  & phone &  &   &  &  & &    &    &    & board &  \\ \hline
    \# shapes & & 2690 & 76 & 55 & 898 & 3758 & 69 & 787 & 392 & 1547 & 451 & 202 & 184 & 283 & 66 & 152 & 5271 \\ \hline
    Wu~\cite{Wu2014248} &  -  & 63.2  & - &      -    & -   & 73.5 & - &    - &    - &     74.4  & -    & - &   -   &   -   &   - & -  &  74.8 \\
    Yi~\cite{Yi16} & 81.4 & 81.0 & 78.4 & 77.7 & \textbf{75.7} & 87.6 & 61.9 & \textbf{92.0} & 85.4 & \textbf{82.5} & \textbf{95.7} & \textbf{70.6} & 91.9 & \textbf{85.9} & 53.1 & 69.8 & 75.3 \\ \hline
    3DCNN & 79.4 & 75.1 & 72.8 & 73.3 & 70.0 & 87.2 & 63.5 & 88.4 & 79.6 & 74.4 & 93.9 & 58.7 & 91.8 & 76.4 & 51.2 & 65.3 & 77.1 \\ 
    Ours & \textbf{83.7} & \textbf{83.4} & \textbf{78.7} & \textbf{82.5} & 74.9 & \textbf{89.6} & \textbf{73.0} & 91.5 & \textbf{85.9} & 80.8 & 95.3 & 65.2 & \textbf{93.0} & 81.2 & \textbf{57.9} & \textbf{72.8} & \textbf{80.6} \\ \hline
    \end{tabular}
    \caption{\textbf{Segmentation results on ShapeNet part dataset.} Metric is mIoU(\%) on points. We compare with two traditional methods \cite{Wu2014248} and \cite{Yi16} and a 3D fully convolutional network baseline proposed by us. Our PointNet method achieved the state-of-the-art in mIoU.}
    \label{tab:segmentation}
\end{table*}

\subsection{Applications}
\label{sec:application}
In this section we show how our network can be trained to perform 3D object classification, object part segmentation and semantic scene segmentation  \footnote{More application examples such as correspondence and point cloud based CAD model retrieval are included in supplementary material.}. Even though we are working on a brand new data representation (point sets), we are able to achieve comparable or even better performance on benchmarks for several tasks.

\paragraph{3D Object Classification} Our network learns global point cloud feature that can be used for object classification. We evaluate our model on the ModelNet40~\cite{wu20153d} shape classification benchmark. There are 12,311 CAD models from 40 man-made object categories, split into 9,843 for training and 2,468 for testing. While previous methods focus on volumetric and mult-view image representations, we are the first to directly work on raw point cloud.

We uniformly sample 1024 points on mesh faces according to face area and normalize them into a unit sphere. During training we augment the point cloud on-the-fly by randomly rotating the object along the up-axis and jitter the position of each points by a Gaussian noise with zero mean and 0.02 standard deviation.

In Table~\ref{tab:classification}, we compare our model with previous works as well as our baseline using MLP on traditional features extracted from point cloud (point density, D2, shape contour etc.).
Our model achieved state-of-the-art performance among methods based on 3D input (volumetric and point cloud). With only fully connected layers and max pooling, our net gains a strong lead in inference speed and can be easily parallelized in CPU as well. There is still a small gap between our method and multi-view based method (MVCNN~\cite{su15mvcnn}), which we think is due to the loss of fine geometry details that can be captured by rendered images.

\paragraph{3D Object Part Segmentation} Part segmentation is a challenging fine-grained 3D recognition task. Given a 3D scan or a mesh model, the task is to assign part category label (e.g. chair leg, cup handle) to each point or face.


We evaluate on ShapeNet part data set from~\cite{Yi16}, which contains 16,881 shapes from 16 categories, annotated with 50 parts in total. Most object categories are labeled with two to five parts. Ground truth annotations are labeled on sampled points on the shapes.

We formulate part segmentation as a per-point classification problem. Evaluation metric is mIoU on points. For each shape S of category C, to calculate the shape's mIoU: For each part type in category C, compute IoU between groundtruth and prediction. If the union of groundtruth and prediction points is empty, then count part IoU as 1. Then we average IoUs for all part types in category C to get mIoU for that shape. To calculate mIoU for the category, we take average of mIoUs for all shapes in that category. 

In this section, we compare our segmentation version PointNet (a modified version of Fig~\ref{fig:pointnet_arch}, \textit{Segmentation Network}) with two traditional methods \cite{Wu2014248} and \cite{Yi16} that both take advantage of point-wise geometry features and correspondences between shapes, as well as our own 3D CNN baseline.
See supplementary for the detailed modifications and network architecture for the 3D CNN.

In Table~\ref{tab:segmentation}, we report per-category and mean IoU(\%) scores. We observe a 2.3\% mean IoU improvement and our net beats the baseline methods in most categories.

\begin{table}[b!]
    \centering
    \small
    \begin{tabular}[width=\linewidth]{l|c|c}
    \hline
    ~             & mean IoU  & overall accuracy \\ \hline
    Ours baseline          &  20.12 & 53.19    \\ \hline
    Ours PointNet          & \textbf{47.71} & \textbf{78.62}  \\ \hline
    \end{tabular}
    \caption{\textbf{Results on semantic segmentation in scenes.} Metric is average IoU over 13 classes (structural and furniture elements plus clutter) and classification accuracy calculated on points. }
    \label{tab:semantic_segmentation}
\end{table}

\begin{table}[b!]
    \centering
    \small
    \begin{tabular}[width=\linewidth]{l|cccc|c}
    \hline
    ~                              & table & chair & sofa & board & mean  \\ \hline
    \# instance & 455 & 1363 & 55 & 137 & ~ \\ \hline
    Armeni et al.~\cite{armeni_cvpr16}          & 46.02 & 16.15 & \textbf{6.78} & 3.91  & 18.22 \\ \hline
    Ours & \textbf{46.67}     & \textbf{33.80 }    & 4.76    & \textbf{11.72}     & \textbf{24.24}     \\ \hline
    \end{tabular}
    \caption{\textbf{Results on 3D object detection in scenes.} Metric is average precision with threshold IoU 0.5 computed in 3D volumes.}
    \label{tab:3d_detection}
\end{table}


We also perform experiments on simulated Kinect scans to test the robustness of these methods. For every CAD model in the ShapeNet part data set, we use Blensor Kinect Simulator~\cite{Gschwandtner11b} to generate incomplete point clouds from six random viewpoints. We train our PointNet on the complete shapes and partial scans with the same network architecture and training setting. Results show that we lose only 5.3\% mean IoU. In Fig~\ref{fig:qualitative_part_segmentation}, we present qualitative results on both complete and partial data. One can see that though partial data is fairly challenging, our predictions are reasonable.

\paragraph{Semantic Segmentation in Scenes} Our network on part segmentation can be easily extended to semantic scene segmentation, where point labels become semantic object classes instead of object part labels.



We experiment on the Stanford 3D semantic parsing data set~\cite{armeni_cvpr16}. The dataset contains 3D scans from Matterport scanners in 6 areas including 271 rooms. Each point in the scan is annotated with one of the semantic labels from 13 categories (chair, table, floor, wall etc. plus clutter).

To prepare training data, we firstly split points by room, and then sample rooms into blocks with area 1m by 1m. We train our segmentation version of PointNet to predict per point class in each block. Each point is represented by a 9-dim vector of XYZ, RGB and normalized location as to the room (from 0 to 1). At training time, we randomly sample 4096 points in each block on-the-fly. At test time, we test on all the points. We follow the same protocol as~\cite{armeni_cvpr16} to use k-fold strategy for train and test.

\begin{figure}[t!]
    \centering
    \includegraphics[width=0.8\linewidth,height=4cm]{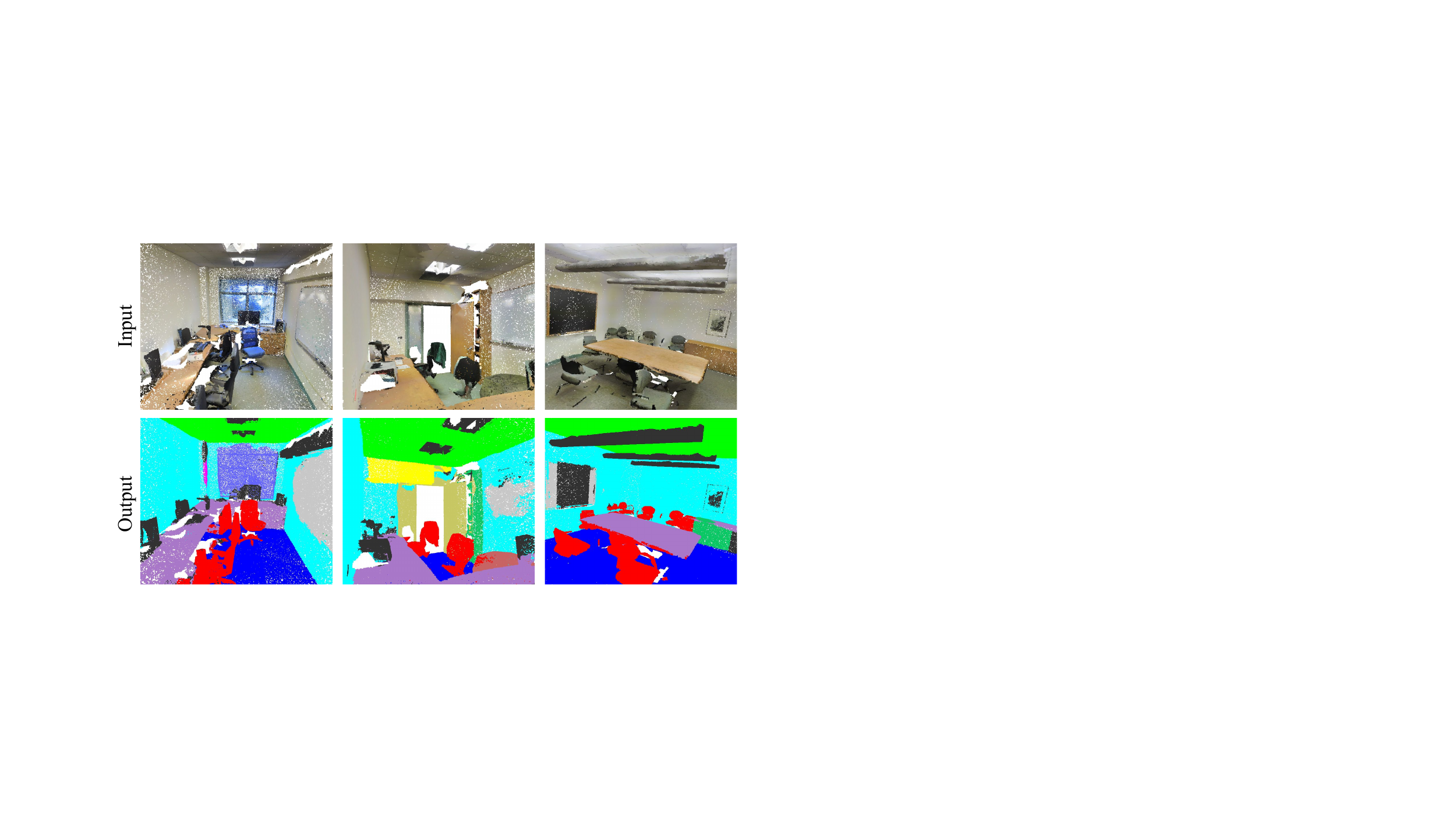}
    \caption{\textbf{Qualitative results for semantic segmentation.} Top row is input point cloud with color. Bottom row is output semantic segmentation result (on points) displayed in the same camera viewpoint as input.}
    \label{fig:qualitative_segmentation}
\end{figure}

We compare our method with a baseline using handcrafted point features. The baseline extracts the same 9-dim local features and three additional ones: local point density, local curvature and normal. We use standard MLP as the classifier.  Results are shown in Table~\ref{tab:semantic_segmentation}, where our PointNet method significantly outperforms the baseline method. In Fig~\ref{fig:qualitative_segmentation}, we show qualitative segmentation results. Our network is able to output smooth predictions and is robust to missing points and occlusions.

Based on the semantic segmentation output from our network, we further build a 3D object detection system using connected component for object proposal (see supplementary for details). We compare with previous state-of-the-art method in Table~\ref{tab:3d_detection}. The previous method is based on a sliding shape method (with CRF post processing) with SVMs trained on local geometric features and global room context feature in voxel grids. Our method outperforms it by a large margin on the furniture categories reported.


\subsection{Architecture Design Analysis}
\label{sec:arch_analysis}

In this section we validate our design choices 
by control experiments. We also show the effects of our network's hyperparameters.

\paragraph{Comparison with Alternative Order-invariant Methods} As mentioned in Sec~\ref{sec:pointnet_arch}, there are at least three options for consuming unordered set inputs. We use the ModelNet40 shape classification problem as a test bed for comparisons of those options, the following two control experiment will also use this task. 

\begin{figure}[t!]
    \centering
    \includegraphics[width=\linewidth]{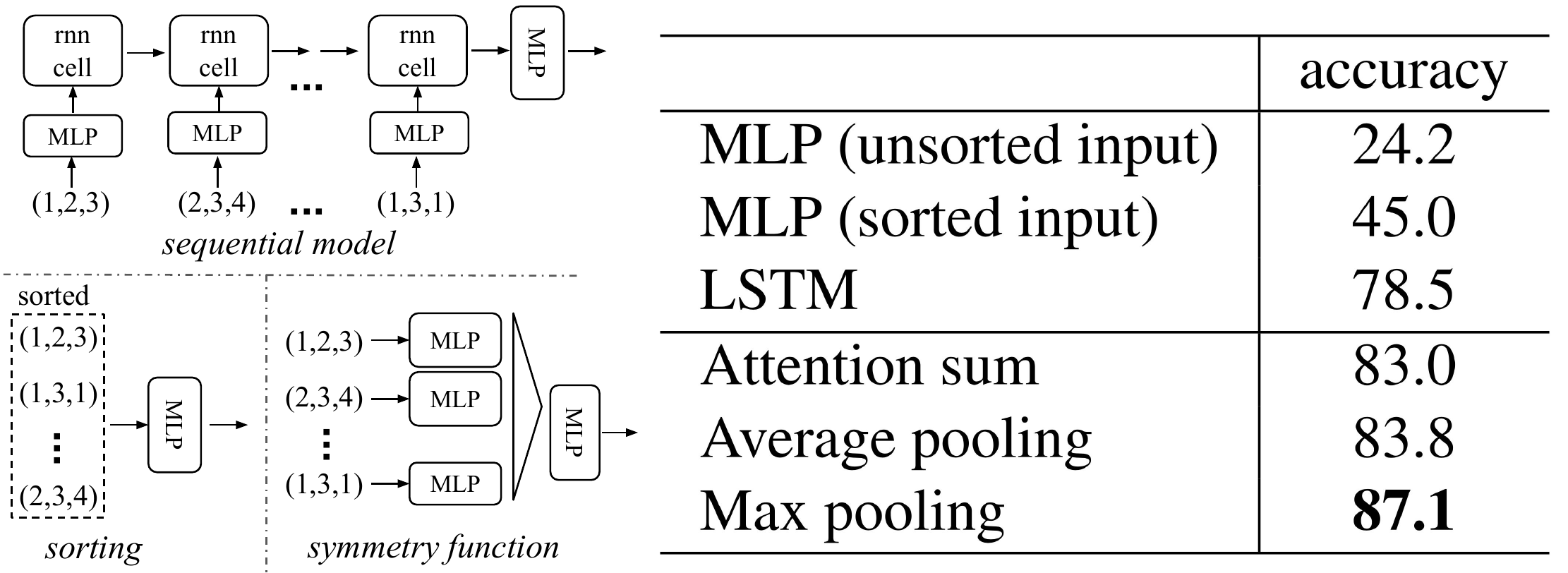}
    \caption{\textbf{Three approaches to achieve order invariance.} Multi-layer perceptron (MLP) applied on points consists of 5 hidden layers with neuron sizes 64,64,64,128,1024, all points share a single copy of MLP. The MLP close to the output consists of two layers with sizes 512,256.
    }
    \label{fig:order_invariant}
\end{figure}

The baselines (illustrated in Fig~\ref{fig:order_invariant}) we compared with include multi-layer perceptron on unsorted and sorted points as $n \times 3$ arrays, RNN model that considers input point as a sequence, and a model based on symmetry functions. The symmetry operation we experimented include max pooling, average pooling and an attention based weighted sum. The attention method is similar to that in~\cite{vinyals2015order}, where a scalar score is predicted from each point feature, then the score is normalized across points by computing a softmax. The weighted sum is then computed on the normalized scores and the point features. As shown in Fig~\ref{fig:order_invariant}, max-pooling operation achieves the best performance by a large winning margin, which validates our choice.


\paragraph{Effectiveness of Input and Feature Transformations} In Table~\ref{tab:transform} we demonstrate the positive effects of our input and feature transformations (for alignment). It's interesting to see that the most basic architecture already achieves quite reasonable results. Using input transformation gives a $0.8\%$ performance boost. The regularization loss is necessary for the higher dimension transform to work. By combining both transformations and the regularization term, we achieve the best performance.

\begin{table}[b!]
    \small
    \centering
    \begin{tabular}[width=\linewidth]{l|c}
    \hline
    Transform              & accuracy \\ \hline
    none                   & 87.1     \\ \hline
    input (3x3)            & 87.9     \\
    feature (64x64)        & 86.9     \\
    feature (64x64) + reg. & 87.4     \\ \hline
    both                   & \textbf{89.2}     \\ \hline
    \end{tabular}
    \caption{\textbf{Effects of input feature transforms.} Metric is overall classification accuracy on ModelNet40 test set.}
    \label{tab:transform}
\end{table}



\begin{figure}
    \centering
    \includegraphics[width=\linewidth]{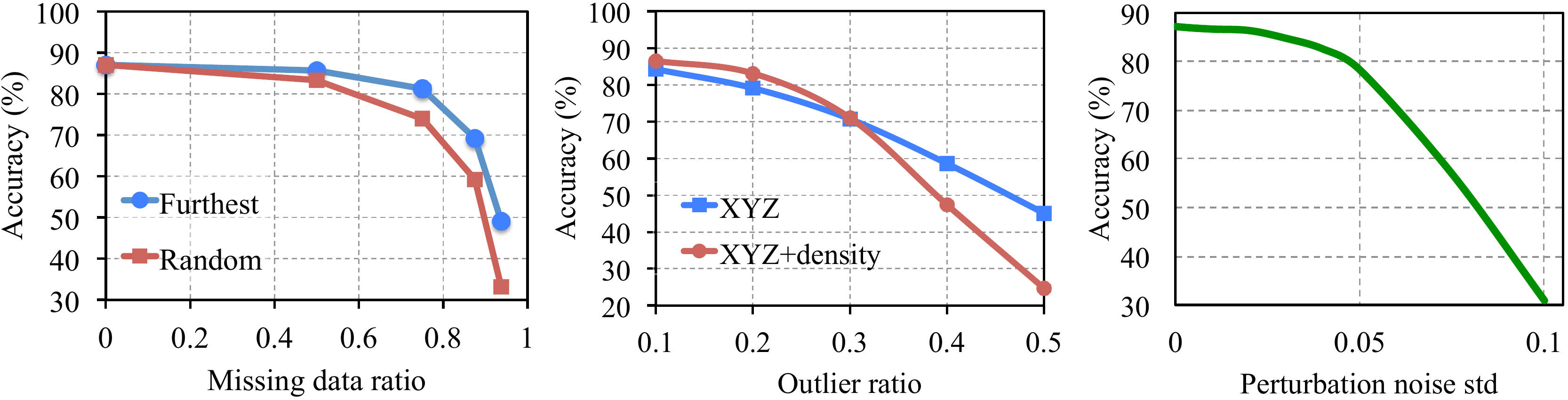}
    \caption{\textbf{PointNet robustness test.} The metric is overall classification accuracy on ModelNet40 test set. Left: Delete points. Furthest means the original 1024 points are sampled with furthest sampling. Middle: Insertion. Outliers uniformly scattered in the unit sphere. Right: Perturbation. Add Gaussian noise to each point independently.}
    \label{fig:robustness}
\end{figure}

\paragraph{Robustness Test} We show our PointNet, while simple and effective, is robust to various kinds of input corruptions. We use the same architecture as in Fig~\ref{fig:order_invariant}'s max pooling network. Input points are normalized into a unit sphere. Results are in Fig~\ref{fig:robustness}.

As to missing points, when there are $50\%$ points missing, the accuracy only drops by $2.4\%$ and $3.8\%$ w.r.t. furthest and random input sampling. Our net is also robust to outlier points, if it has seen those during training. We evaluate two models: one trained on points with $(x,y,z)$ coordinates; the other on $(x,y,z)$ plus point density. The net has more than $80\%$ accuracy even when $20\%$ of the points are outliers. Fig~\ref{fig:robustness} right shows the net is robust to point perturbations.

\subsection{Visualizing PointNet}
\label{sec:visualizing_pointnet}

\begin{figure}[b]
    \centering
    \includegraphics[width=0.8\linewidth]{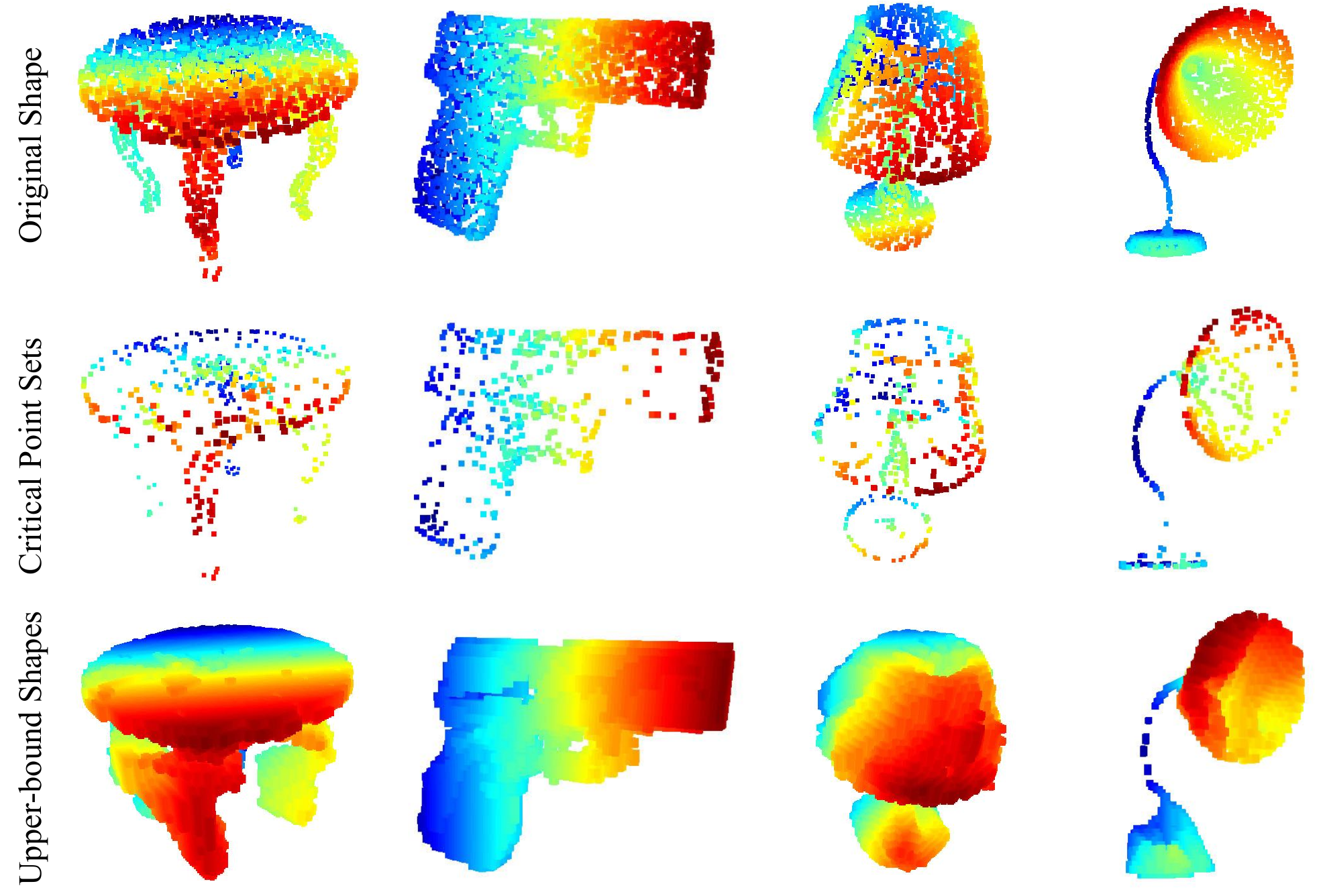}
    \caption{\textbf{Critical points and upper bound shape.} While critical points jointly determine the global shape feature for a given shape, any point cloud that falls between the critical points set and the upper bound shape gives exactly the same feature. We color-code all figures to show the depth information. }
    \label{fig:recon}
\end{figure}

In Fig~\ref{fig:recon}, we visualize \textit{critical point sets} $\mathcal{C}_S$ and \textit{upper-bound shapes} $\mathcal{N}_S$ (as discussed in Thm~\ref{thm:thm2}) for some sample shapes $S$. The point sets between the two shapes will give exactly the same global shape feature $f(S)$. 

We can see clearly from Fig~\ref{fig:recon} that the \textit{critical point sets} $\mathcal{C}_S$, those contributed to the max pooled feature, summarizes the skeleton of the shape.
The \textit{upper-bound shapes} $\mathcal{N}_S$ illustrates the largest possible point cloud that give the same global shape feature $f(S)$ as the input point cloud $S$. $\mathcal{C}_S$ and $\mathcal{N}_S$ reflect the robustness of PointNet, meaning that losing some non-critical points does not change the global shape signature $f(S)$ at all.


The $\mathcal{N}_S$ is constructed by forwarding all the points in a edge-length-2 cube through the network and select points $p$ whose point function values $(h_1(p), h_2(p), \cdots, h_K(p))$ are no larger than the global shape descriptor. 

\subsection{Time and Space Complexity Analysis}
\label{sec:complexity}
Table~\ref{pointnet_complexity} summarizes space (number of parameters in the network) and time (floating-point operations/sample) complexity of our classification PointNet. We also compare PointNet to a representative set of volumetric and multi-view based architectures in previous works.

While MVCNN~\cite{su15mvcnn} and Subvolume (3D CNN) ~\cite{qi2016volumetric} achieve high performance, PointNet is orders more efficient in computational cost (measured in FLOPs/sample: \emph{141x} and \emph{8x} more efficient, respectively). Besides, PointNet is much more space efficient than MVCNN in terms of \#param in the network (\emph{17x} less parameters).
Moreover, PointNet is much more scalable -- it's space and time complexity is $O(N)$ -- \emph{linear} in the number of input points. However, since convolution dominates computing time, multi-view method's time complexity grows \emph{squarely} on image resolution and volumetric convolution based method grows \emph{cubically} with the volume size.

Empirically, PointNet is able to process more than one million points per second for point cloud classification (around 1K objects/second) or semantic segmentation (around 2 rooms/second) with a 1080X GPU on TensorFlow, showing great potential for real-time applications. 

\begin{table}[h!]
    \centering
    \begin{tabular}{|l|l|l|l|}
    \hline
    ~                & \#params & FLOPs/sample\\ \hline
    PointNet (vanilla)  & 0.8M                        & 148M \\
    PointNet         & 3.5M                         & 440M \\ \hline
    Subvolume~\cite{qi2016volumetric} & 16.6M                        & 3633M \\ \hline
    MVCNN~\cite{su15mvcnn}   & 60.0M                          & 62057M \\ \hline
    \end{tabular}
    \caption{\textbf{Time and space complexity of deep architectures for 3D data classification.} PointNet (vanilla) is the classification PointNet without input and feature transformations. FLOP stands for floating-point operation.
    The ``M'' stands for million. Subvolume and MVCNN used pooling on input data from multiple rotations or views, without which they have much inferior performance.}
    \label{pointnet_complexity}
    \vspace{-3mm}
\end{table}

\section{Conclusion}
\label{sec:conclusion}
In this work, we propose a novel deep neural network \emph{PointNet} that directly consumes point cloud. Our network provides a unified approach to a number of 3D recognition tasks including object classification, part segmentation and semantic segmentation, while obtaining on par or better results than state of the arts on standard benchmarks. We also provide theoretical analysis and visualizations towards understanding of our network.

\mypara{Acknowledgement.} The authors gratefully acknowledge the support of a Samsung GRO grant, ONR MURI N00014-13-1-0341 grant, NSF grant IIS-1528025, a Google Focused Research Award, a gift from the Adobe corporation and hardware donations by NVIDIA.

{\small
\bibliographystyle{ieee}
\bibliography{pcl}
}

\newpage
\appendix
\section*{Supplementary}

\section{Overview}
This document provides additional quantitative results, technical details and more qualitative test examples to the main paper.

In Sec~\ref{sec:cla_robust} we extend the robustness test to compare PointNet with VoxNet on incomplete input. In Sec~\ref{sec:network} we provide more details on neural network architectures, training parameters and in Sec~\ref{sec:detection} we describe our detection pipeline in scenes. Then Sec~\ref{sec:supp_application} illustrates more applications of PointNet, while Sec~\ref{sec:architecture} shows more analysis experiments. Sec~\ref{sec:proof} provides a proof for our theory on PointNet. At last, we show more visualization results in Sec~\ref{sec:visu}.

\section{Comparison between PointNet and VoxNet (Sec 5.2)}
\label{sec:cla_robust}
We extend the experiments in Sec 5.2 Robustness Test to compare PointNet and VoxNet~\cite{maturana2015voxnet} (a representative architecture for volumetric representation) on robustness to missing data in the input point cloud. Both networks are trained on the same train test split with 1024 number of points as input. For VoxNet we voxelize the point cloud to $32 \times 32 \times 32$ occupancy grids and augment the training data by random rotation around up-axis and jittering.

\begin{figure}[h!]
    \centering
    \includegraphics[width=0.7\linewidth]{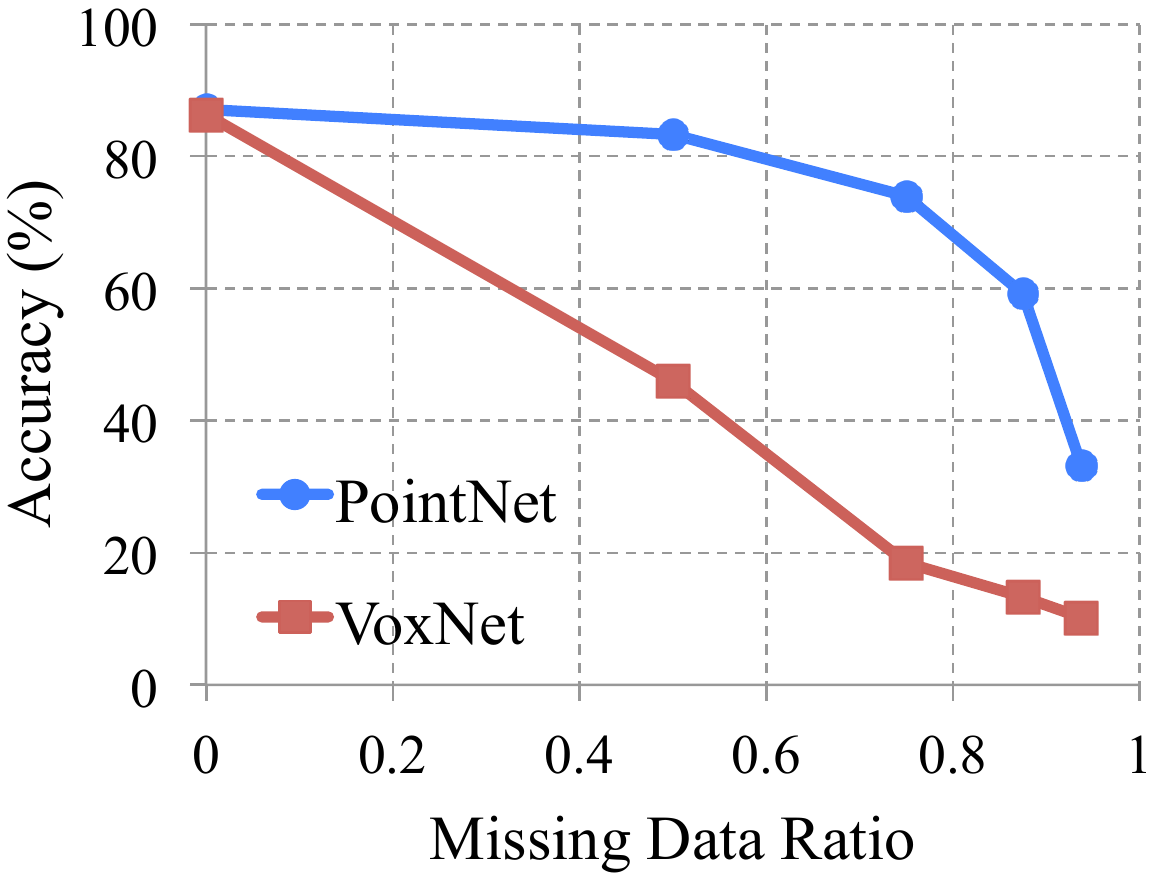}
    \caption{\textbf{PointNet v.s. VoxNet~\cite{maturana2015voxnet} on incomplete input data.} Metric is overall classification accurcacy on ModelNet40 test set. Note that VoxNet is using 12 viewpoints averaging while PointNet is using only one view of the point cloud. Evidently PointNet presents much stronger robustness to missing points.}
    \label{fig:compare}
\end{figure}

At test time, input points are randomly dropped out by a certain ratio. As VoxNet is sensitive to rotations, its prediction uses average scores from 12 viewpoints of a point cloud. As shown in Fig~\ref{fig:compare}, we see that our PointNet is much more robust to missing points. VoxNet's accuracy dramatically drops when half of the input points are missing, from $86.3\%$ to $46.0\%$ with a $40.3\%$ difference, while our PointNet only has a $3.7\%$ performance drop. This can be explained by the theoretical analysis and explanation of our PointNet -- it is learning to use a collection of \textit{critical points} to summarize the shape, thus it is very robust to missing data.

\section{Network Architecture and Training Details (Sec 5.1)}
\label{sec:network}
\paragraph{PointNet Classification Network} As the basic architecture is already illustrated in the main paper, here we provides more details on the joint alignment/transformation network and training parameters.

The first transformation network is a mini-PointNet that takes raw point cloud as input and regresses to a $3\times3$ matrix. It's composed of a shared $MLP(64,128,1024)$ network (with layer output sizes 64, 128, 1024) on each point, a max pooling across points and two fully connected layers with output sizes $512$, $256$. The output matrix is initialized as an identity matrix. All layers, except the last one, include ReLU and batch normalization. The second transformation network has the same architecture as the first one except that the output is a  $64\times64$ matrix. The matrix is also initialized as an identity. A regularization loss (with weight 0.001) is added to the softmax classification loss to make the matrix close to orthogonal.

We use dropout with keep ratio $0.7$ on the last fully connected layer, whose output dimension $256$, before class score prediction. The decay rate for batch normalization starts with $0.5$ and is gradually increased to $0.99$. We use adam optimizer with initial learning rate $0.001$, momentum $0.9$ and batch size $32$. The learning rate is divided by 2 every 20 epochs. Training on ModelNet takes 3-6 hours to converge with TensorFlow and a GTX1080 GPU.

\paragraph{PointNet Segmentation Network} The segmentation network is an extension to the classification PointNet. Local point features (the output after the second transformation network) and global feature (output of the max pooling) are concatenated for each point. No dropout is used for segmentation network. Training parameters are the same as the classification network.

\begin{figure}
\centering
\includegraphics[width=\linewidth]{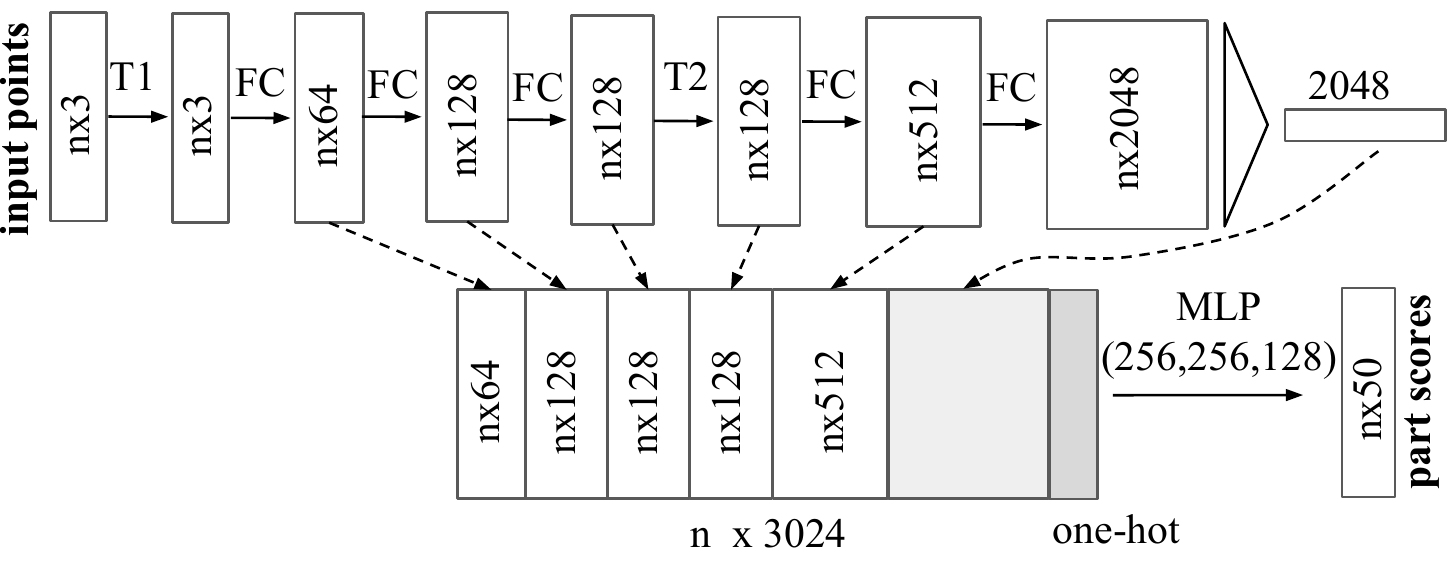}
\caption{\textbf{Network architecture for part segmentation.} T1 and T2 are alignment/transformation networks for input points and features. FC is fully connected layer operating on each point. MLP is multi-layer perceptron on each point. One-hot is a vector of size 16 indicating category of the input shape.}
\label{fig:part_seg_net}
\end{figure}

As to the task of shape part segmentation, we made a few modifications to the basic segmentation network architecture (Fig 2 in main paper) in order to achieve best performance, as illustrated in Fig~\ref{fig:part_seg_net}. We add a one-hot vector indicating the class of the input and concatenate it with the max pooling layer's output. We also increase neurons in some layers and add skip links to collect local point features in different layers and concatenate them to form point feature input to the segmentation network.

While \cite{Wu2014248} and \cite{Yi16} deal with each object category independently, due to the lack of training data for some categories (the total number of shapes for all the categories in the data set are shown in the first line), we train our PointNet across categories (but with one-hot vector input to indicate category). To allow fair comparison, when testing these two models, we only predict part labels for the given specific object category. 

As to semantic segmentation task, we used the architecture as in Fig 2 in the main paper.

It takes around six to twelve hours to train the model on ShapeNet part dataset and around half a day to train on the Stanford semantic parsing dataset.

\begin{figure}[t!]
\centering
\includegraphics[width=\linewidth]{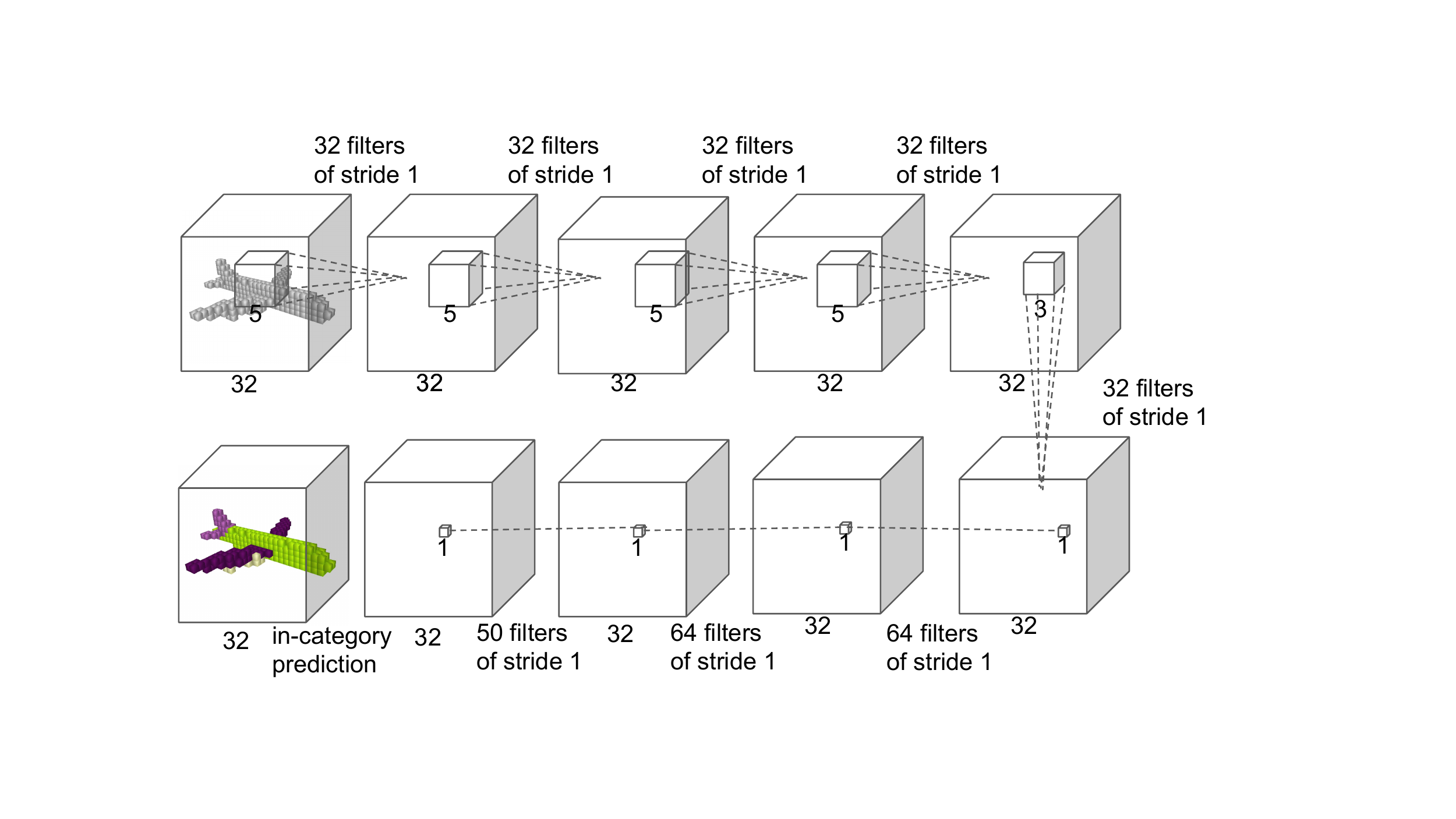}
\caption{\textbf{Baseline 3D CNN segmentation network.} The network is fully convolutional and predicts part scores for each voxel.}
\label{fig:voxnet}
\end{figure}

\paragraph{Baseline 3D CNN Segmentation Network}
In ShapeNet part segmentation experiment, we compare our proposed segmentation version PointNet to two traditional methods as well as a 3D volumetric CNN network baseline. In Fig~\ref{fig:voxnet}, we show the baseline 3D volumetric CNN network we use. We generalize the well-known 3D CNN architectures, such as VoxNet \cite{maturana2015voxnet} and 3DShapeNets \cite{wu20153d} to a fully convolutional 3D CNN segmentation network.


For a given point cloud, we first convert it to the volumetric representation as a occupancy grid with resolution $32 \times 32 \times 32$. Then, five 3D convolution operations each with 32 output channels and stride of 1 are sequentially applied to extract features. The receptive field is 19 for each voxel. 
Finally, a sequence of 3D convolutional layers with kernel size $1\times 1\times 1$ is appended to the computed feature map to predict segmentation label for each voxel. ReLU and batch normalization are used for all layers except the last one. The network is trained across categories, however, in order to compare with other baseline methods where object category is given, we only consider output scores in the given object category.

\section{Details on Detection Pipeline (Sec 5.1)}
\label{sec:detection}
We build a simple 3D object detection system based on the semantic segmentation results and our object classification PointNet.

We use connected component with segmentation scores to get object proposals in scenes. Starting from a random point in the scene, we find its predicted label and use BFS to search nearby points with the same label, with a search radius of $0.2$ meter. If the resulted cluster has more than 200 points (assuming a 4096 point sample in a 1m by 1m area), the cluster's bounding box is marked as one object proposal. For each proposed object, it's detection score is computed as the average point score for that category. Before evaluation, proposals with extremely small areas/volumes are pruned. For tables, chairs and sofas, the bounding boxes are extended to the floor in case the legs are separated with the seat/surface.

We observe that in some rooms such as auditoriums lots of objects (e.g. chairs) are close to each other, where connected component would fail to correctly segment out individual ones. Therefore we leverage our classification network and uses sliding shape method to alleviate the problem for the chair class. We train a binary classification network for each category and use the classifier for sliding window detection. The resulted boxes are pruned by non-maximum suppression. The proposed boxes from connected component and sliding shapes are combined for final evaluation.

\begin{figure}
 \includegraphics[width=0.8\linewidth]{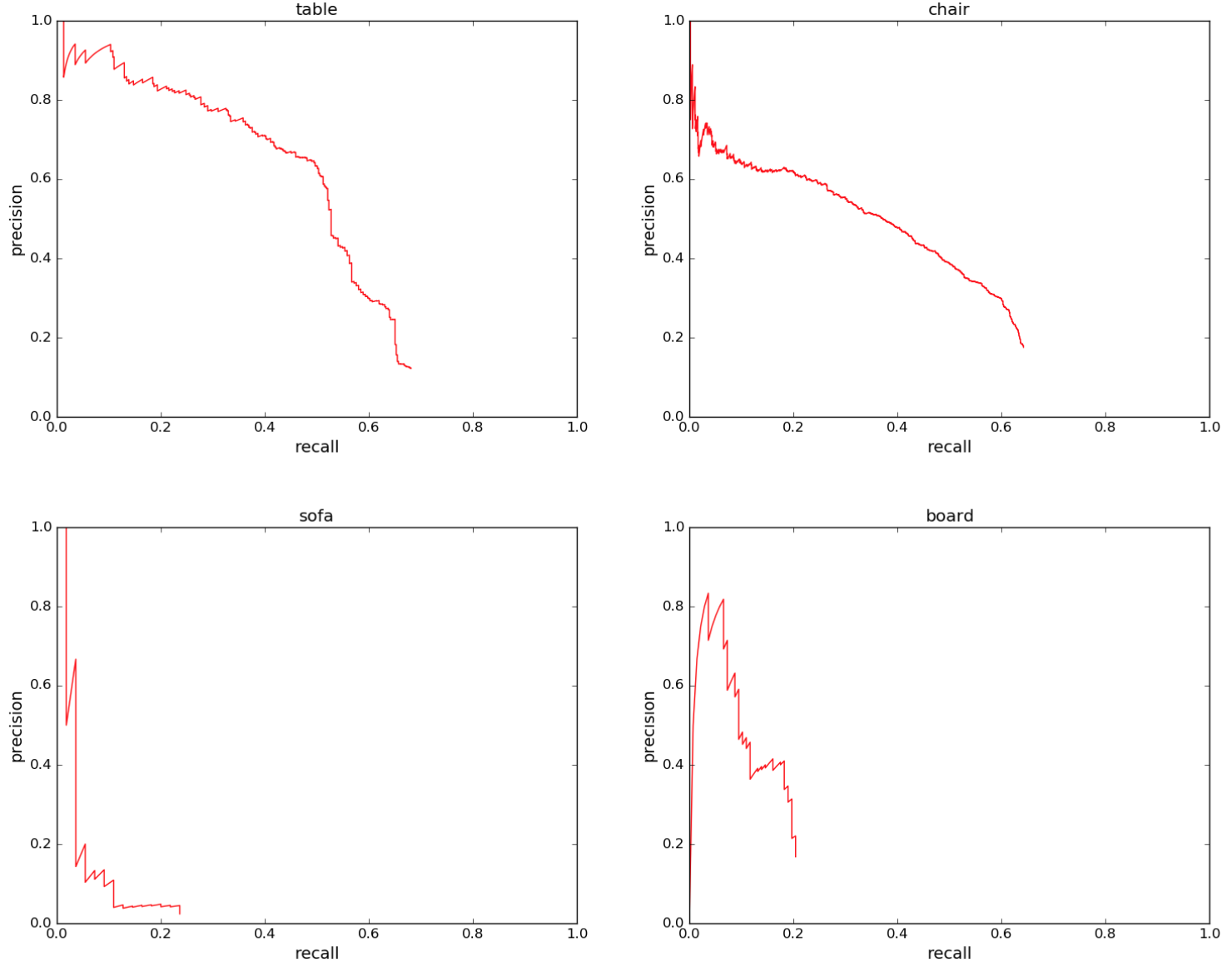}
 \centering
 \caption{\textbf{Precision-recall curves for object detection in 3D point cloud.} We evaluated on all six areas for four categories: table, chair, sofa and board. IoU threshold is 0.5 in volume.}
 \label{fig:pr_curve}
 \end{figure}

 In Fig~\ref{fig:pr_curve}, we show the precision-recall curves for object detection. We trained six models, where each one of them is trained on five areas and tested on the left area. At test phase, each model is tested on the area it has never seen. The test results for all six areas are aggregated for the PR curve generation.
 
\section{More Applications (Sec 5.1)}
\label{sec:supp_application}
\begin{figure}[h]
    \centering
    \includegraphics[width=\linewidth]{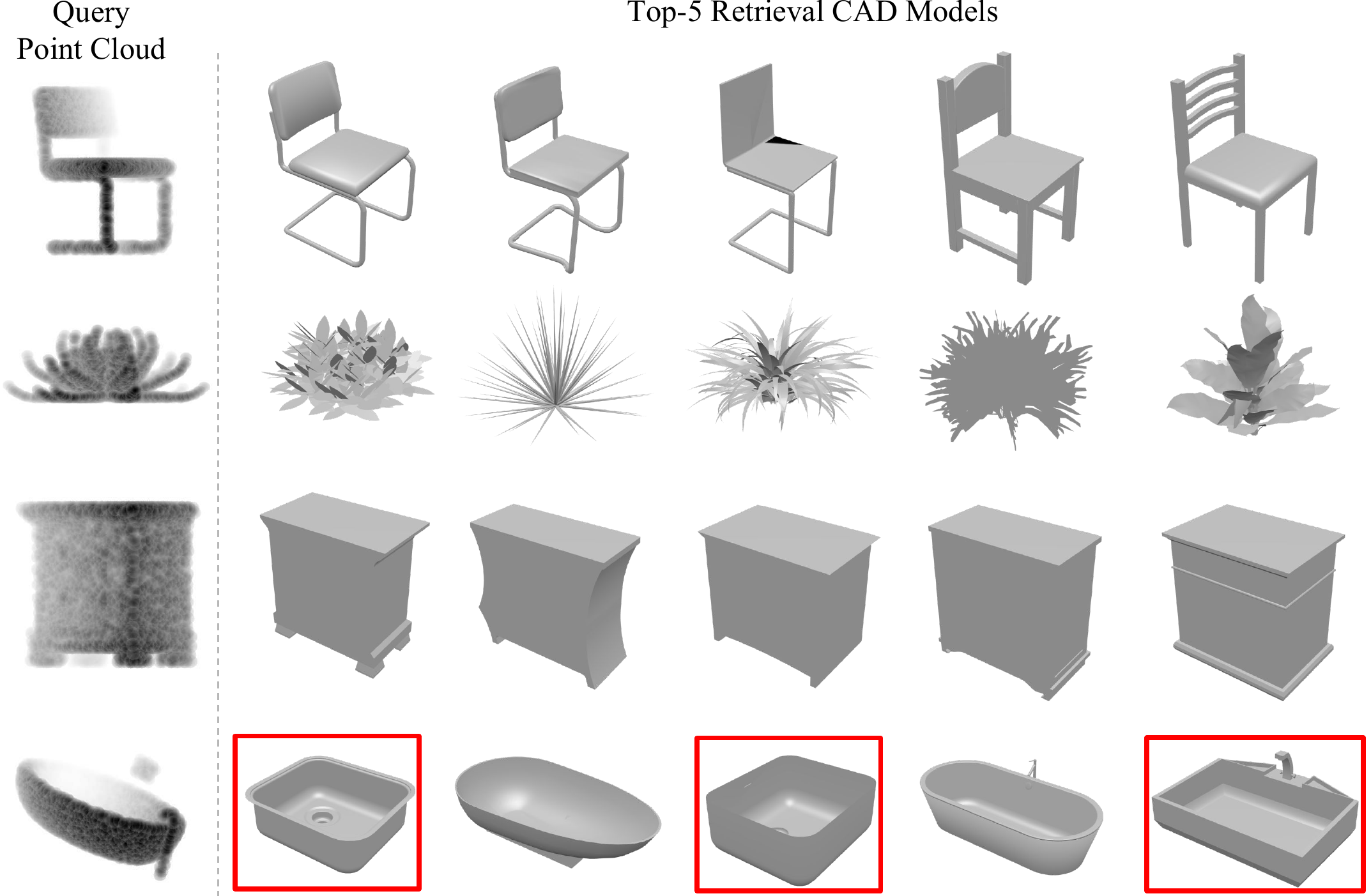}
    \caption{\textbf{Model retrieval from point cloud.} For every given point cloud, we retrieve the top-5 similar shapes from the ModelNet test split. From top to bottom rows, we show examples of chair, plant, nightstand and bathtub queries. Retrieved results that are in wrong category are marked by red boxes.}
    \label{fig:retrieval}
\end{figure}

\paragraph{Model Retrieval from Point Cloud} Our PointNet learns a global shape signature for every given input point cloud. We expect geometrically similar shapes have similar global signature. In this section, we test our conjecture on the shape retrieval application. To be more specific, for every given query shape from ModelNet test split, we compute its global signature (output of the layer before the score prediction layer) given by our classification PointNet and retrieve similar shapes in the train split by nearest neighbor search. Results are shown in Fig~\ref{fig:retrieval}.

\paragraph{Shape Correspondence}

\begin{figure}[h]
    \centering
    \includegraphics[width=\linewidth]{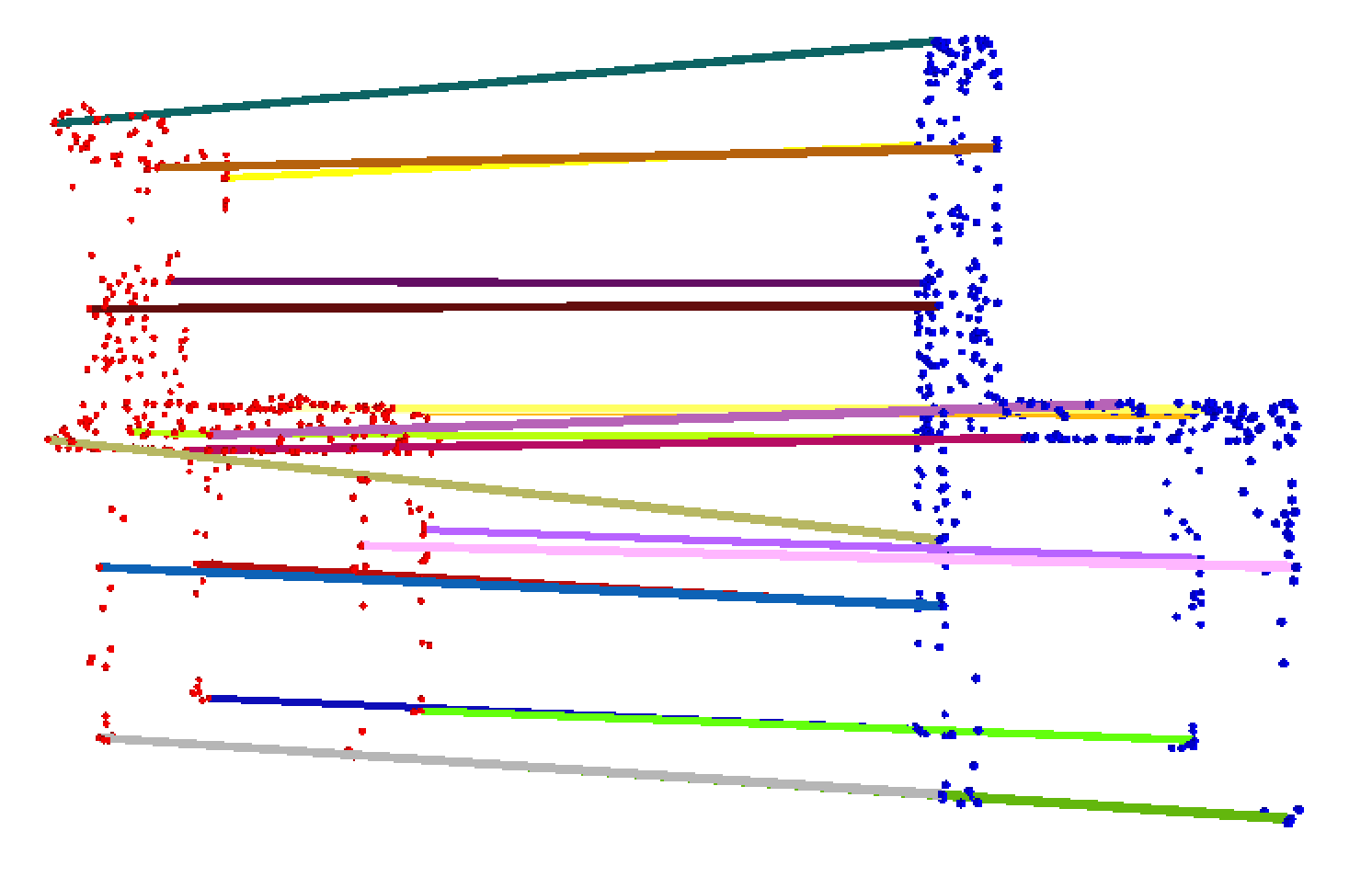}
    \caption{\textbf{Shape correspondence between two chairs.} For the clarity of the visualization, we only show 20 randomly picked correspondence pairs.}
    \label{fig:chair_corr}
\end{figure}

\begin{figure}[h]
    \centering
    \includegraphics[width=\linewidth]{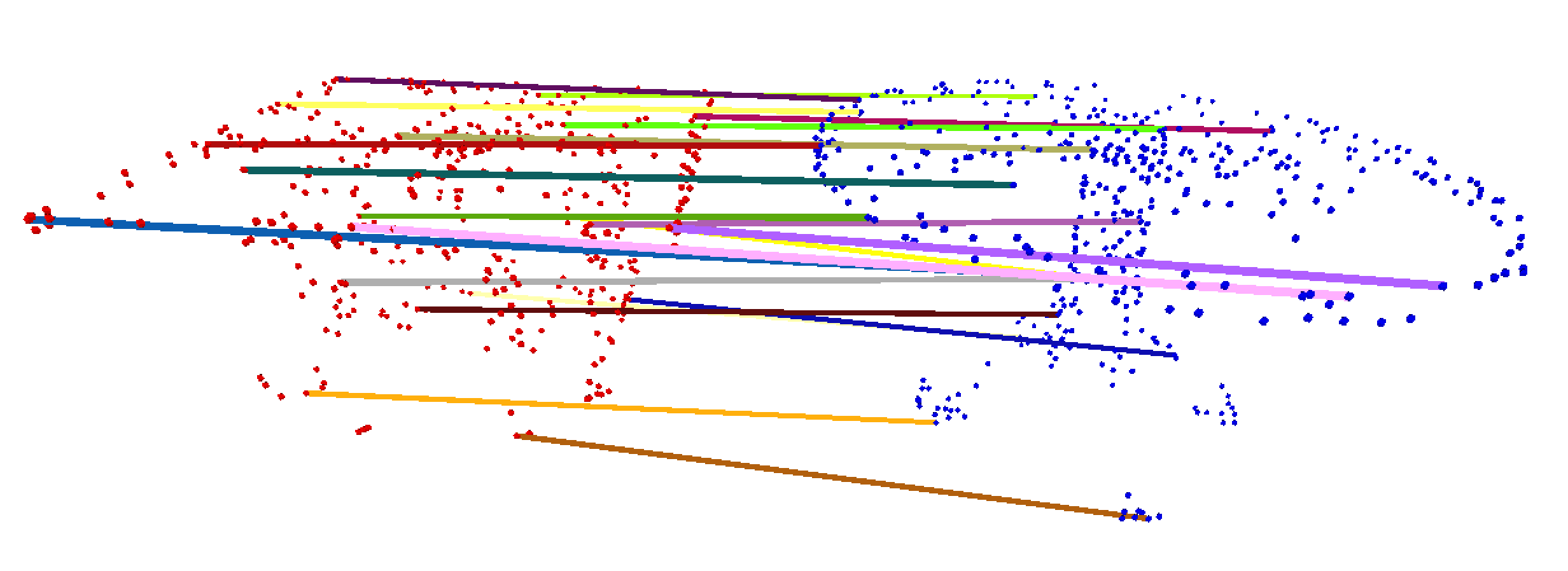}
    \caption{\textbf{Shape correspondence between two tables.} For the clarity of the visualization, we only show 20 randomly picked correspondence pairs.}
    \label{fig:table_corr}
\end{figure}

In this section, we show that point features learnt by PointNet can be potentially used to compute shape correspondences. Given two shapes, we compute the correspondence between their \textit{critical point sets} $C_S$'s by matching the pairs of points that activate the same dimensions in the global features. Fig~\ref{fig:chair_corr} and Fig~\ref{fig:table_corr} show the detected shape correspondence between two similar chairs and tables.

\section{More Architecture Analysis (Sec 5.2)}
\label{sec:architecture}

\begin{figure}[h]
    \centering
    \includegraphics[width=0.8\linewidth]{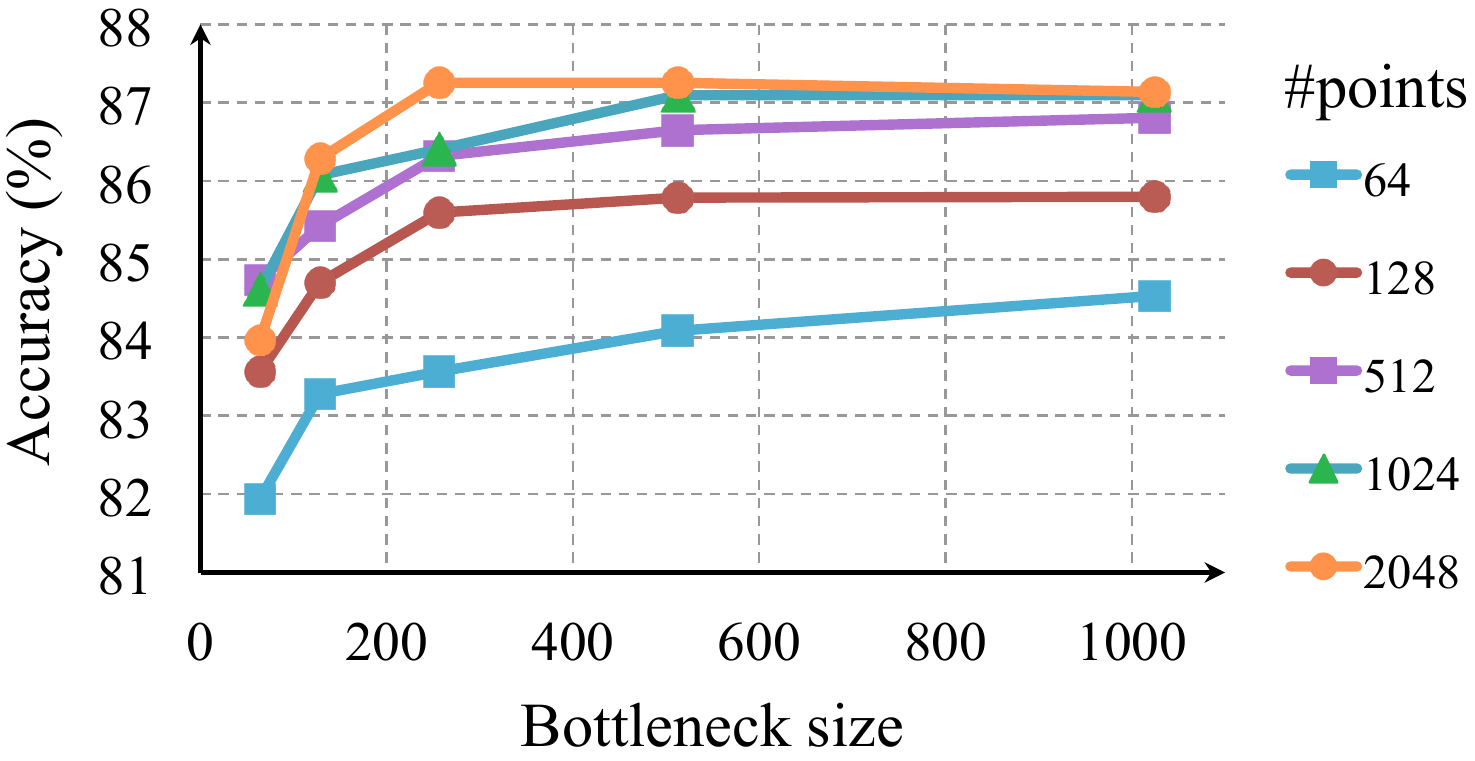}
    \caption{\textbf{Effects of bottleneck size and number of input points.} The metric is overall classification accuracy on ModelNet40 test set.}
    \label{fig:net_param}
\end{figure}

\paragraph{Effects of Bottleneck Dimension and Number of Input Points}
Here we show our model's performance change with regard to the size of the first max layer output as well as the number of input points. In Fig~\ref{fig:net_param} we see that performance grows as we increase the number of points however it saturates at around 1K points. The max layer size plays an important role, increasing the layer size from 64 to 1024 results in a $2-4\%$ performance gain. It indicates that we need enough point feature functions to cover the 3D space in order to discriminate different shapes.

It's worth notice that even with 64 points as input (obtained from furthest point sampling on meshes), our network can achieve decent performance.

\paragraph{MNIST Digit Classification}
While we focus on 3D point cloud learning, a sanity check experiment is to apply our network on a 2D point clouds - pixel sets.

To convert an MNIST image into a 2D point set we threshold pixel values and add the pixel (represented as a point with $(x,y)$ coordinate in the image) with values larger than 128 to the set. We use a set size of 256. If there are more than 256 pixels int he set, we randomly sub-sample it; if there are less, we pad the set with the one of the pixels in the set (due to our max operation, which point to use for the padding will not affect outcome).

As seen in Table~\ref{tab:mnist}, we compare with a few baselines including multi-layer perceptron that considers input image as an ordered vector, a RNN that consider input as sequence from pixel (0,0) to pixel (27,27), and a vanilla version CNN. While the best performing model on MNIST is still well engineered CNNs (achieving less than $0.3\%$ error rate), it's interesting to see that our PointNet model can achieve reasonable performance by considering image as a 2D point set.

\begin{table}[h!]
    \centering
    \begin{tabular}[width=\linewidth]{l|c|c}
    \hline
    ~                      & input & error (\%) \\ \hline
    Multi-layer perceptron~\cite{simard2003best} & vector & 1.60  \\
    LeNet5~\cite{lecun1998gradient}                 & image & 0.80 \\ \hline
    Ours PointNet          & point set & 0.78 \\ \hline
    \end{tabular}
    \caption{\textbf{MNIST classification results.} We compare with vanilla versions of other deep architectures to show that our network based on point sets input is achieving reasonable performance on this traditional task.}
    \label{tab:mnist}
\end{table}

\paragraph{Normal Estimation}
In segmentation version of PointNet, local point features and global feature are concatenated in order to provide context to local points. However, it's unclear whether the context is learnt through this concatenation. In this experiment, we validate our design by showing that our segmentation network can be trained to predict point normals, a local geometric property that is determined by a point's neighborhood.


We train a modified version of our segmentation PointNet in a supervised manner to regress to the ground-truth point normals. We just change the last layer of our segmentation PointNet to predict normal vector for each point. We use absolute value of cosine distance as loss.

\begin{figure}[t!]
\centering
\includegraphics[width=0.9\linewidth]{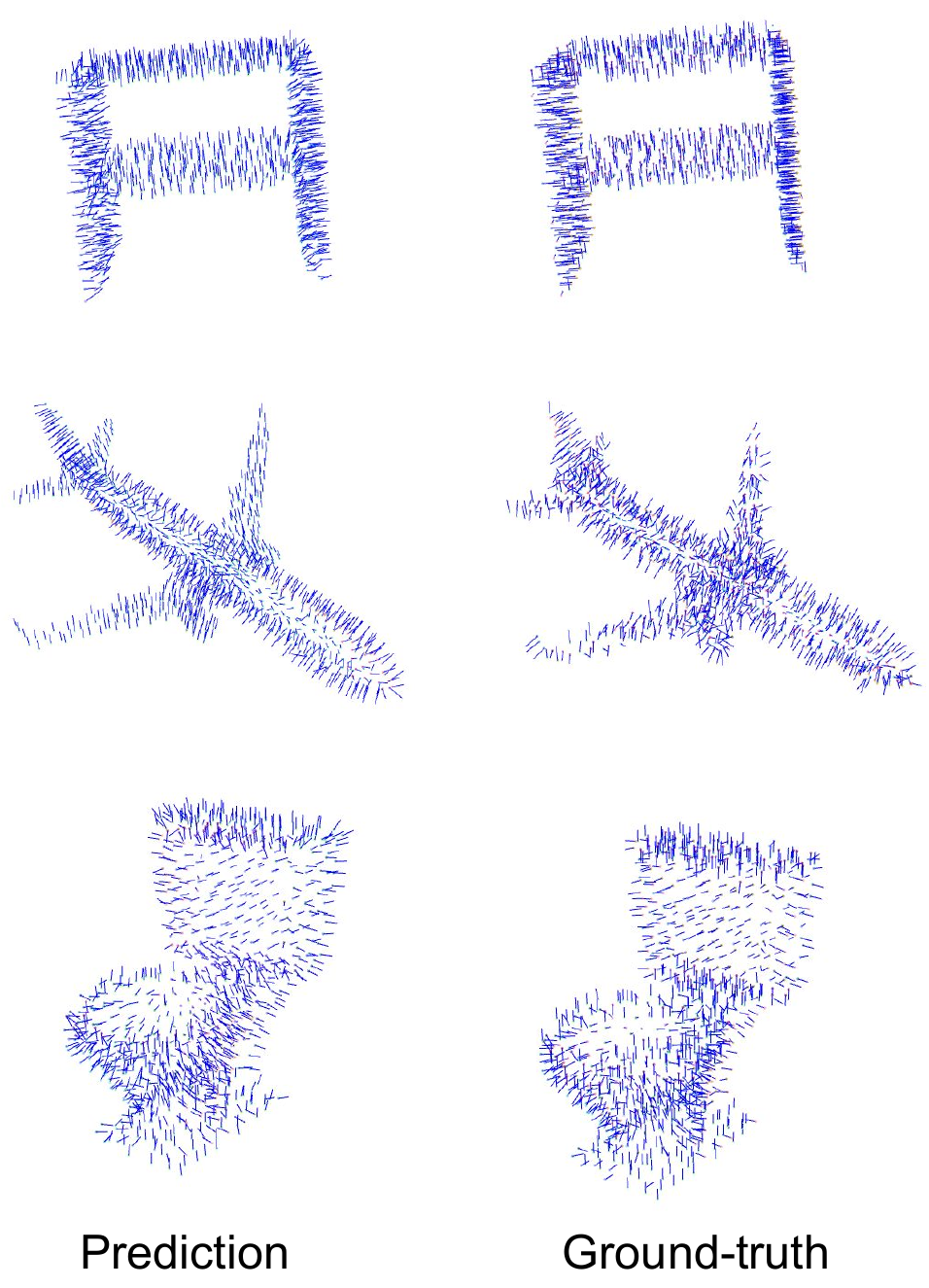}
\caption{\textbf{PointNet normal reconstrution results.} In this figure, we show the reconstructed normals for all the points in some sample point clouds and the ground-truth normals computed on the mesh.}
\label{fig:normal_recon}
\end{figure}

Fig.~\ref{fig:normal_recon} compares our PointNet normal prediction results (the left columns) to the ground-truth normals computed from the mesh (the right columns). We observe a reasonable normal reconstruction. Our predictions are more smooth and continuous than the ground-truth which includes flipped normal directions in some region.

\begin{figure}[t!]
\centering
\includegraphics[width=0.9\linewidth]{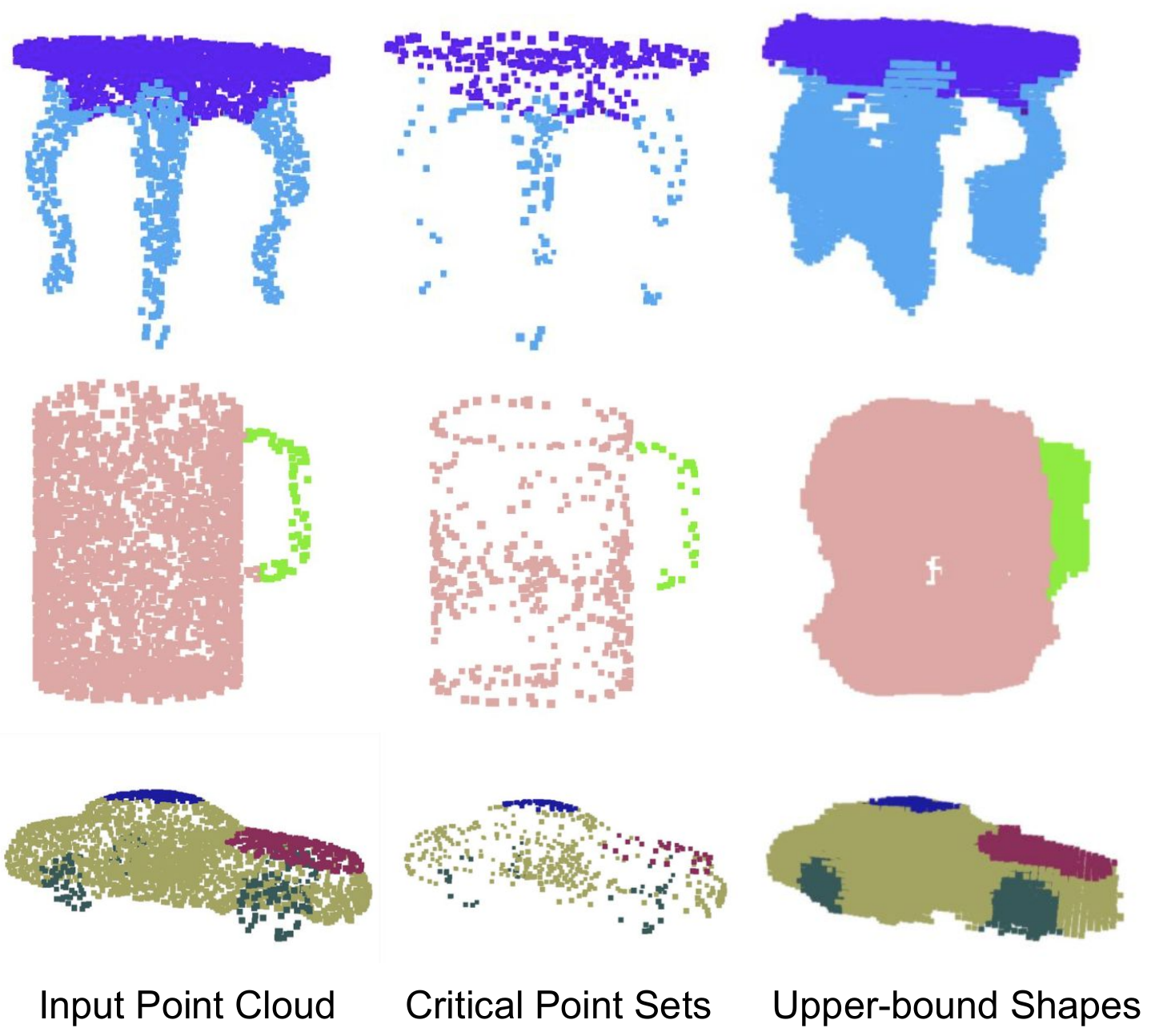}
\caption{\textbf{The consistency of segmentation results.} We illustrate the segmentation results for some sample given point clouds $S$, their \textit{critical point sets} $\mathcal{C}_S$ and \textit{upper-bound shapes} $\mathcal{N}_S$. We observe that the shape family between the $\mathcal{C}_S$ and $\mathcal{N}_S$ share a consistent segmentation results.}
\label{fig:seg_robust}
\end{figure}

\paragraph{Segmentation Robustness} As discussed in Sec 5.2 and Sec~\ref{sec:cla_robust}, our PointNet is less sensitive to data corruption and missing points for classification tasks since the global shape feature is extracted from a collection of \textit{critical points} from the given input point cloud. In this section, we show that the robustness holds for segmentation tasks too. The per-point part labels are predicted based on the combination of per-point features and the learnt global shape feature. In Fig~\ref{fig:seg_robust}, we illustrate the segmentation results for the given input point clouds $S$ (the left-most column), the \textit{critical point sets} $\mathcal{C}_S$ (the middle column) and the \textit{upper-bound shapes} $\mathcal{N}_S$.

\begin{figure}[t!]
\centering
\includegraphics[width=\linewidth]{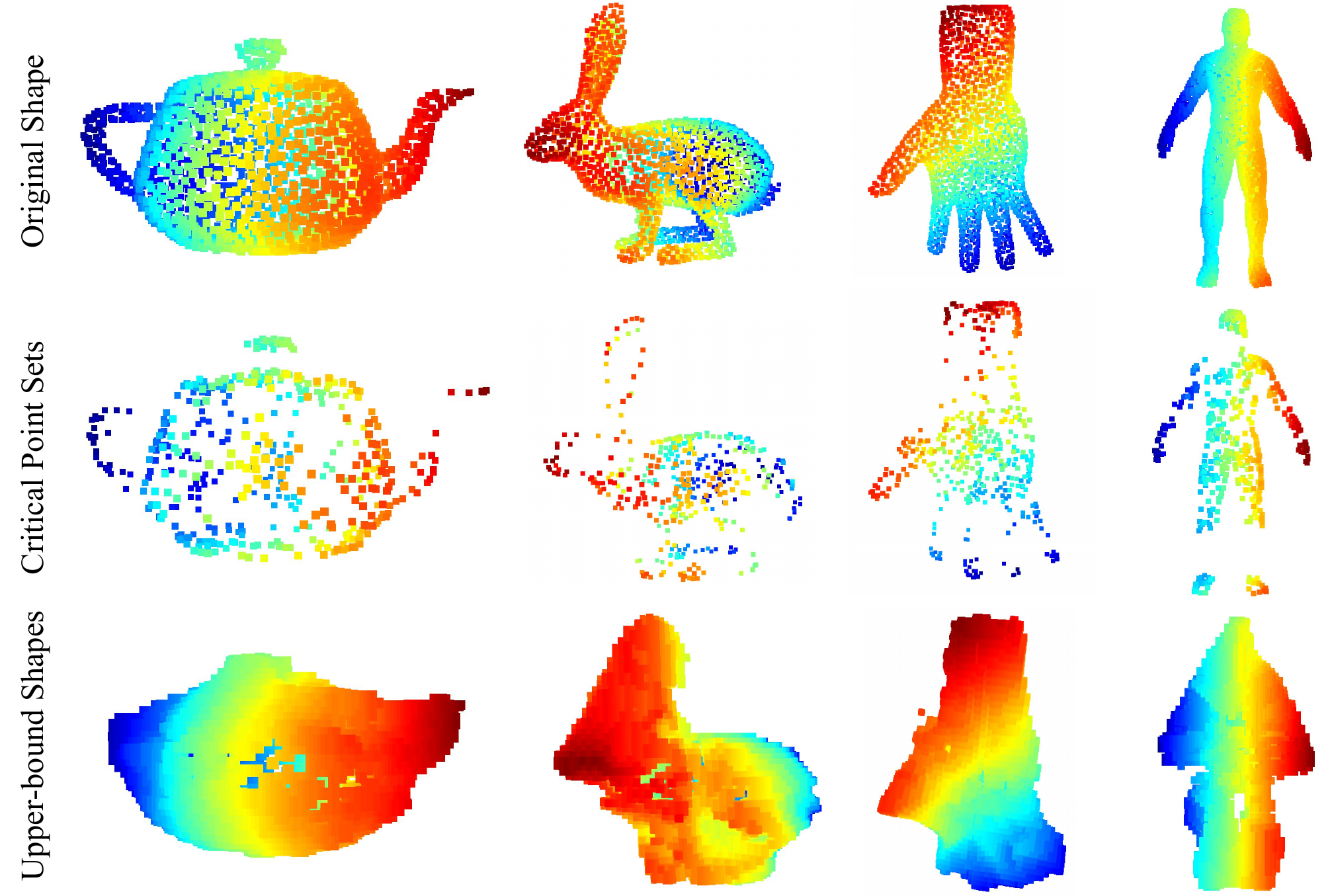}
\caption{\textbf{The critical point sets and the upper-bound shapes for unseen objects.} We visualize the \textit{critical point sets} and the \textit{upper-bound shapes} for teapot, bunny, hand and human body, which are not in the ModelNet or ShapeNet shape repository to test the generalizability of the learnt per-point functions of our PointNet on other unseen objects. The images are color-coded to reflect the depth information.}
\label{fig:unseen}
\end{figure}

\paragraph{Network Generalizability to Unseen Shape Categories}
In Fig~\ref{fig:unseen}, we visualize the \textit{critical point sets} and the \textit{upper-bound shapes} for new shapes from unseen categories (face, house, rabbit, teapot) that are not present in ModelNet or ShapeNet. It shows that the learnt per-point functions are generalizable. However, since we train mostly on man-made objects with lots of planar structures, the reconstructed upper-bound shape in novel categories also contain more planar surfaces. 
 
\section{Proof of Theorem (Sec 4.3)}
\label{sec:proof}
Let $\mathcal{X}=\{S: S\subseteq [0,1]\mbox{ and } |S|=n \}$. 

$f:\mathcal{X}\rightarrow \mathbb{R}$ is a continuous function on $\mathcal{X}$ w.r.t to Hausdorff distance $d_H(\cdot, \cdot)$ if the following condition is satisfied:

$\forall \epsilon > 0, \exists \delta >0$, for any $S, S'\in\mathcal{X}$, if $d_H(S, S') < \delta$, then $|f(S)-f(S')|< \epsilon$.

We show that $f$ can be approximated arbitrarily by composing a symmetric function and a continuous function. 

\begin{customthm}{1}
Suppose $f:\mathcal{X}\rightarrow \mathbb{R}$ is a continuous set function w.r.t Hausdorff distance $d_H(\cdot, \cdot)$. $\forall \epsilon > 0$, $\exists$ a continuous function $h$ and a symmetric function $g(x_1, \dots, x_n)=\gamma \circ \mbox{MAX}$, where $\gamma$ is a continuous function, $\mbox{MAX}$ is a vector max operator that takes $n$ vectors as input and returns a new vector of the element-wise maximum, such that for any $S\in\mathcal{X}$,
\begin{align*}
	|f(S) - \gamma(\mbox{MAX}(h(x_1), \ldots, h(x_n)))| < \epsilon
\end{align*}
where $x_1, \ldots, x_n$ are the elements of $S$ extracted in certain order, 
\end{customthm}

\begin{proof}
By the continuity of $f$, we take $\delta_{\epsilon}$ so that 
$|f(S)-f(S')|<\epsilon$ for any $S, S'\in \mathcal{X} \mbox{ if } d_H(S, S')<\delta_{\epsilon}$. 

Define $K=\lceil 1/\delta_{\epsilon}\rceil$, which split $[0,1]$ into $K$ intervals evenly and define an auxiliary function that maps a point to the left end of the interval it lies in:
$$\sigma(x)=\frac{\lfloor K x \rfloor}{K}$$
Let $\tilde{S}=\{\sigma(x):x\in S\}$, then
$$|f(S)-f(\tilde{S})|< \epsilon$$
because $d_H(S, \tilde{S})<1/K\le \delta_{\epsilon}$.

Let $h_k(x)=e^{-d(x, [\frac{k-1}{K}, \frac{k}{K}])}$ be a soft indicator function where $d(x, I)$ is the point to set (interval) distance. Let $\myvec h(x)=[h_1(x); \ldots; h_K(x)]$, then $\myvec h:\mathbb{R}\rightarrow \mathbb{R}^K$. 

Let $v_j(x_1, \ldots, x_n)=\max\{\tilde{h}_j(x_1),\ldots,\tilde{h}_j(x_n)\}$, indicating the occupancy of the $j$-th interval by points in $S$. Let $\myvec v=[v_1;\ldots; v_K]$, then $\myvec v:\underbrace{\mathbb{R}\times \ldots\times \mathbb{R}}_{n}\rightarrow \{0, 1\}^K$ is a symmetric function, indicating the occupancy of each interval by points in $S$. 

Define $\tau:\{0, 1\}^K\rightarrow \mathcal{X}$ as $\tau(v)=\{\frac{k-1}{K}: v_k\ge 1\}$, which maps the occupancy vector to a set which contains the left end of each occupied interval. It is easy to show:
\begin{align*}
\tau(\myvec v(x_1, \ldots, x_n))\equiv \tilde{S} 
\end{align*}
where $x_1,\ldots,x_n$ are the elements of $S$ extracted in certain order.

Let $\gamma:\mathbb{R}^K\rightarrow \mathbb{R}$ be a continuous function such that $\gamma(\myvec v)=f(\tau(\myvec v))$ for $v\in\{0, 1\}^K$. Then,
\begin{align*}
&|\gamma(\myvec v(x_1, \ldots, x_n))-f(S)|\\
=&|f(\tau(\myvec v(x_1, \ldots, x_n)))-f(S)|<\epsilon
\end{align*}

Note that $\gamma(\myvec v(x_1, \ldots, x_n))$ can be rewritten as follows:
\begin{align*}
\gamma(\myvec v(x_1, \ldots, x_n))=&\gamma(\mbox{MAX}(\myvec h(x_1), \ldots, \myvec h(x_n)))\\
=&(\gamma \circ \mbox{MAX}) (\myvec h(x_1),\ldots,\myvec h(x_n))
\end{align*}
Obviously $\gamma \circ \mbox{MAX}$ is a symmetric function.
\end{proof}

Next we give the proof of Theorem 2.
We define $\myvec u=\underset{x_i\in S}{\mbox{MAX}}\{h(x_i)\}$ to be the sub-network of $f$ which maps a point set in $[0,1]^m$ to a $K$-dimensional vector. The following theorem tells us that small corruptions or extra noise points in the input set is not likely to change the output of our network:
\begin{customthm}{2}
Suppose $\myvec u:\mathcal{X}\rightarrow \mathbb{R}^K$ such that $\myvec u=\underset{x_i\in S}{\mbox{MAX}}\{h(x_i)\}$ and $f=\gamma \circ \myvec u$. Then, 
\begin{enumerate}[label=(\alph*)]   
    \item $\forall S, \exists~\mathcal{C}_S, \mathcal{N}_S\subseteq \mathcal{X}$,  $f(T)=f(S)$ if  $\mathcal{C}_S\subseteq T\subseteq \mathcal{N}_S$;
    \item $|\mathcal{C}_S| \le K$
\end{enumerate}
\end{customthm}
\begin{proof}
Obviously, $\forall S\in \mathcal{X}$, $f(S)$ is determined by $\myvec u(S)$. So we only need to prove that 
$\forall S, \exists\,\mathcal{C}_S, \mathcal{N}_S\subseteq \mathcal{X}, f(T)=f(S)\,\mbox{if}\,\mathcal{C}_S\subseteq T\subseteq \mathcal{N}_S$. 

For the $j$th dimension as the output of $\myvec u$, there exists at least one $x_j \in \mathcal{X}$ such that $h_j(x_j)=\myvec u_j$, where $h_j$ is the $j$th dimension of the output vector from $h$. Take $\mathcal{C}_S$ as the union of all $x_j$ for $j=1,\ldots,K$. Then, $\mathcal{C}_S$ satisfies the above condition. 

Adding any additional points $x$ such that $h(x)\le \myvec u(S)$ at all dimensions to $\mathcal{C}_S$ does not change $\myvec u$, hence $f$. Therefore, $\mathcal{T}_S$ can be obtained adding the union of all such points to $\mathcal{N}_S$.

\end{proof}

\begin{figure}[h!]
    \centering
    \includegraphics[width=0.7\linewidth]{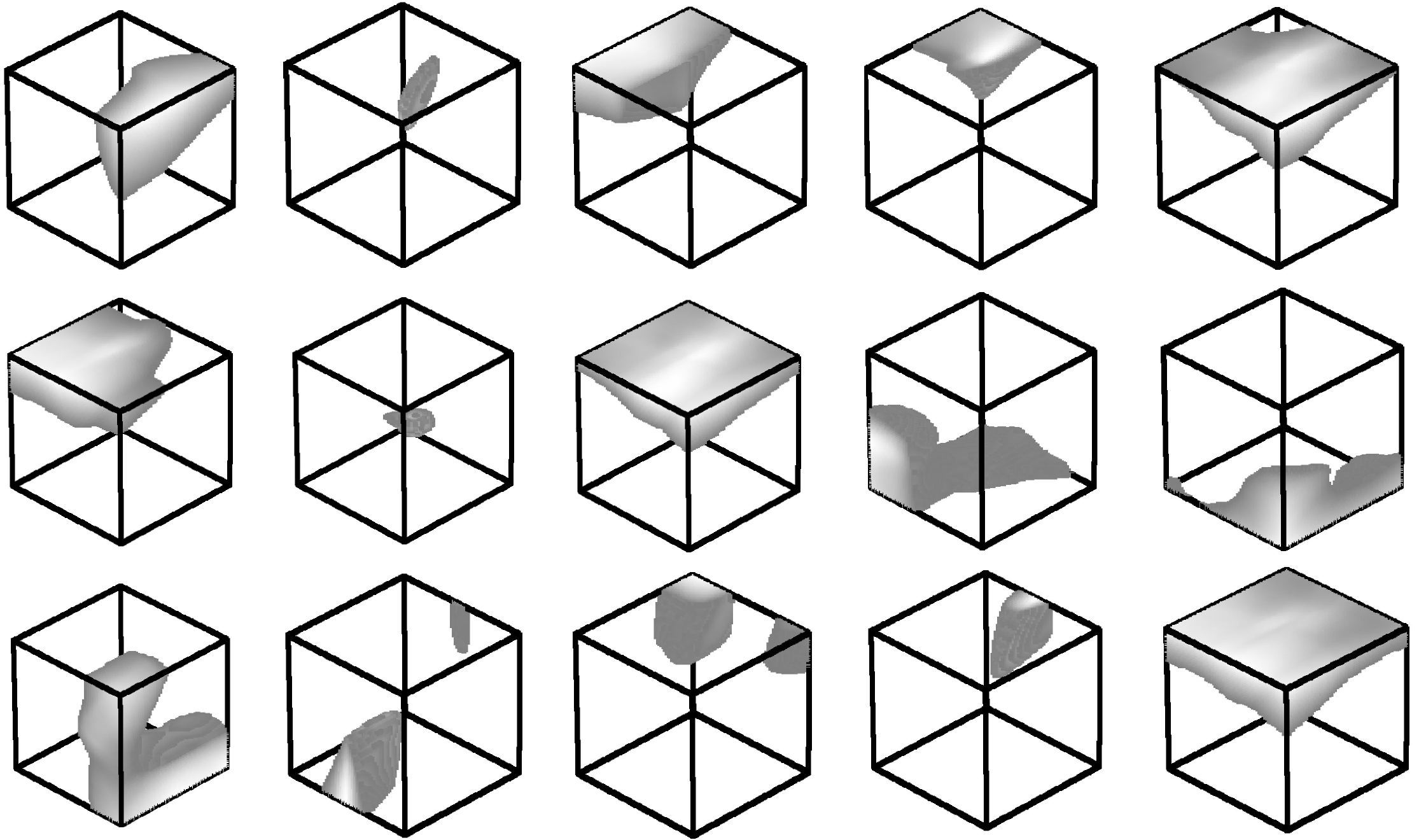}
    \caption{\textbf{Point function visualization.} For each per-point function $h$, we calculate the values $h(p)$ for all the points $p$ in a cube of diameter two located at the origin, which spatially covers the unit sphere to which our input shapes are normalized when training our PointNet. In this figure, we visualize all the points $p$ that give $h(p)>0.5$ with function values color-coded by the brightness of the voxel. We randomly pick 15 point functions and visualize the activation regions for them.}
    \label{fig:functions}
\end{figure}

\section{More Visualizations}
\label{sec:visu}
\paragraph{Classification Visualization}

\begin{figure*}[t!]
\centering
\includegraphics[width=\linewidth]{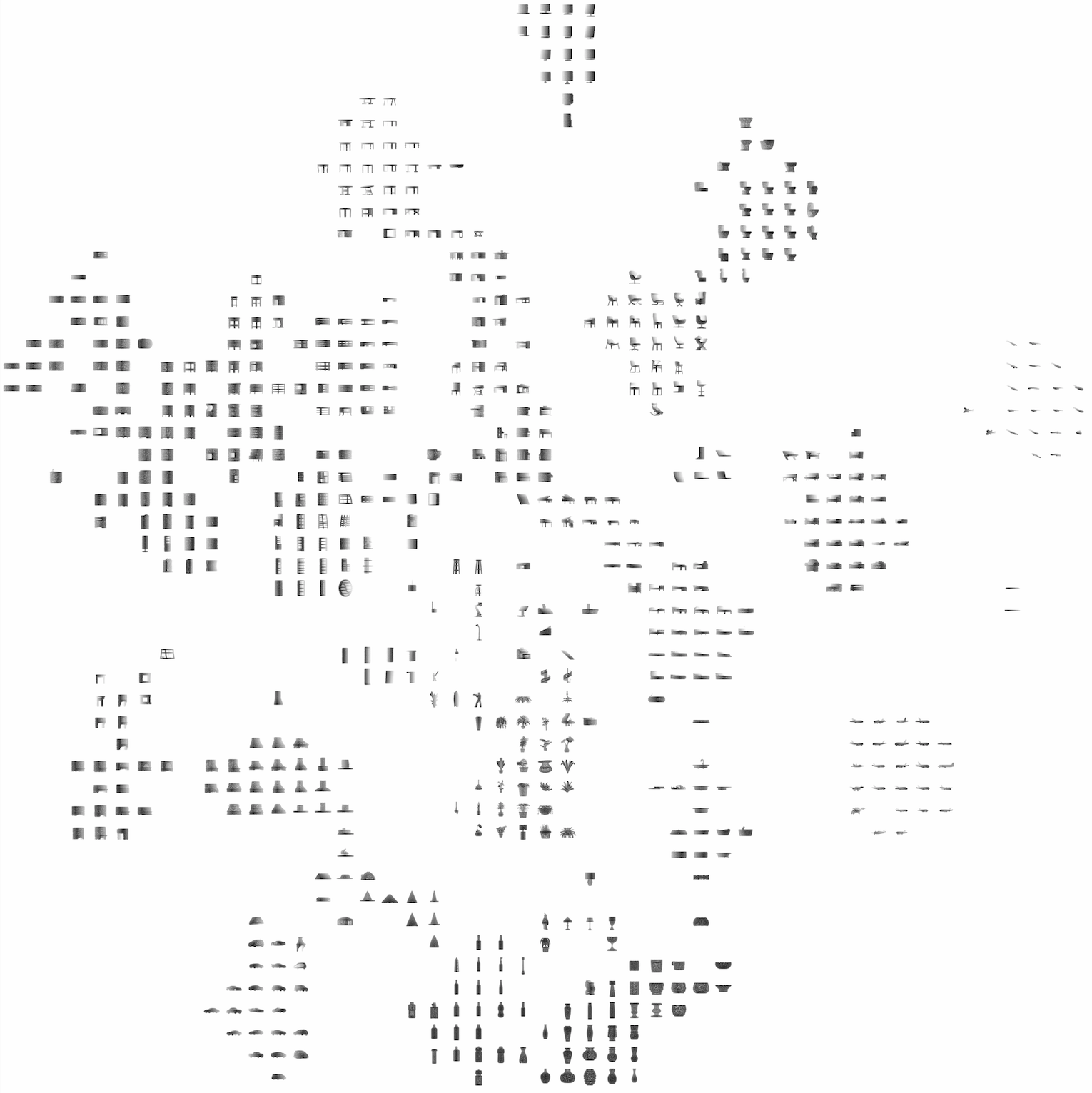}
\caption{\textbf{2D embedding of learnt shape global features.} We use t-SNE technique to visualize the learnt global shape features for the shapes in ModelNet40 test split.}
\label{fig:tsne}
\end{figure*}

We use t-SNE\cite{maaten2008visualizing} to embed point cloud global signature (1024-dim) from our classification PointNet into a 2D space. Fig~\ref{fig:tsne} shows the embedding space of ModelNet 40 test split shapes. Similar shapes are clustered together according to their semantic categories.

\begin{figure*}[t!]
\centering
\includegraphics[width=0.82\linewidth]{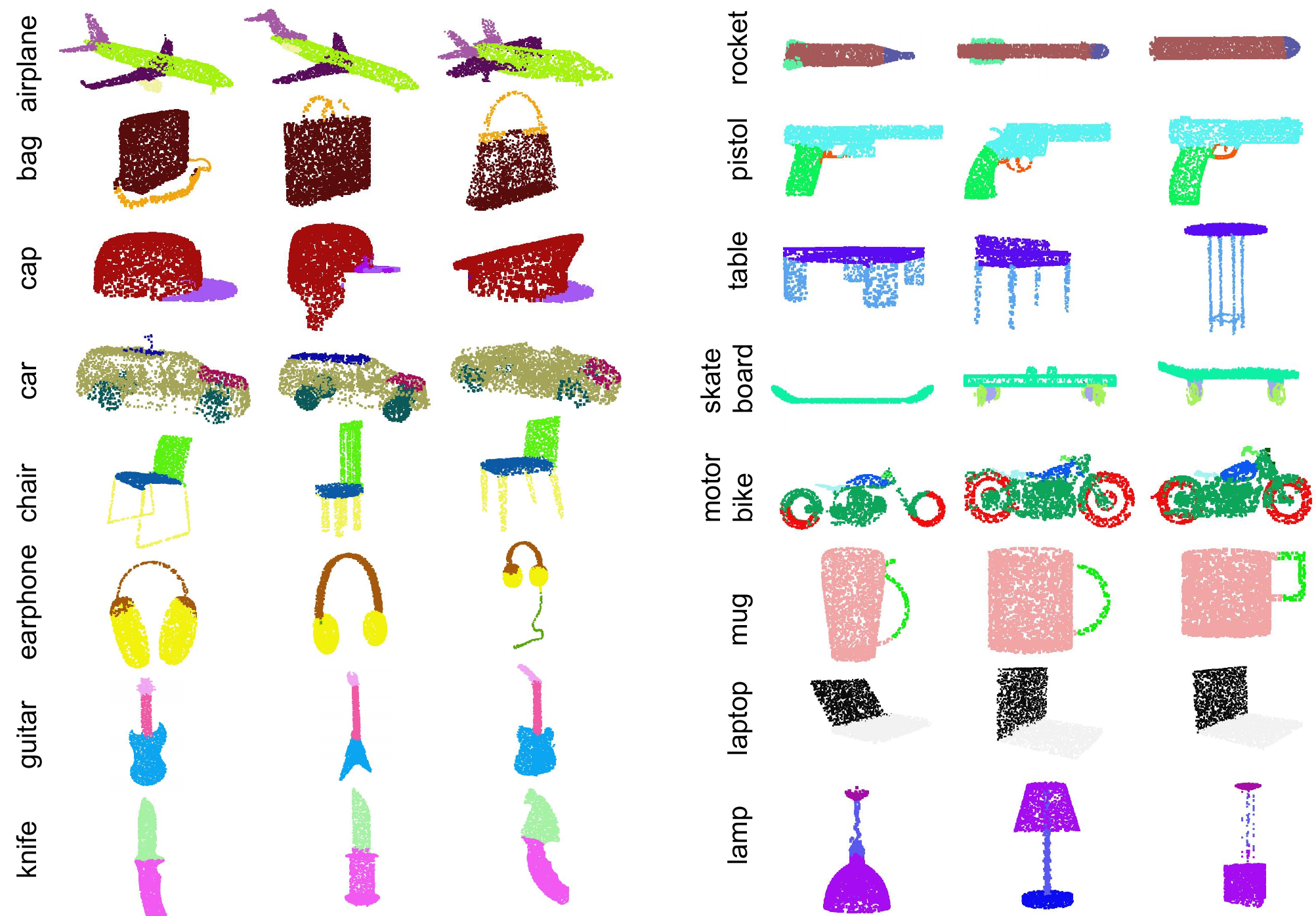}
\caption{\textbf{PointNet segmentation results on complete CAD models.} }
\label{fig:part_seg_complete}
\end{figure*}

\begin{figure*}[t!]
\centering
\includegraphics[width=0.82\linewidth]{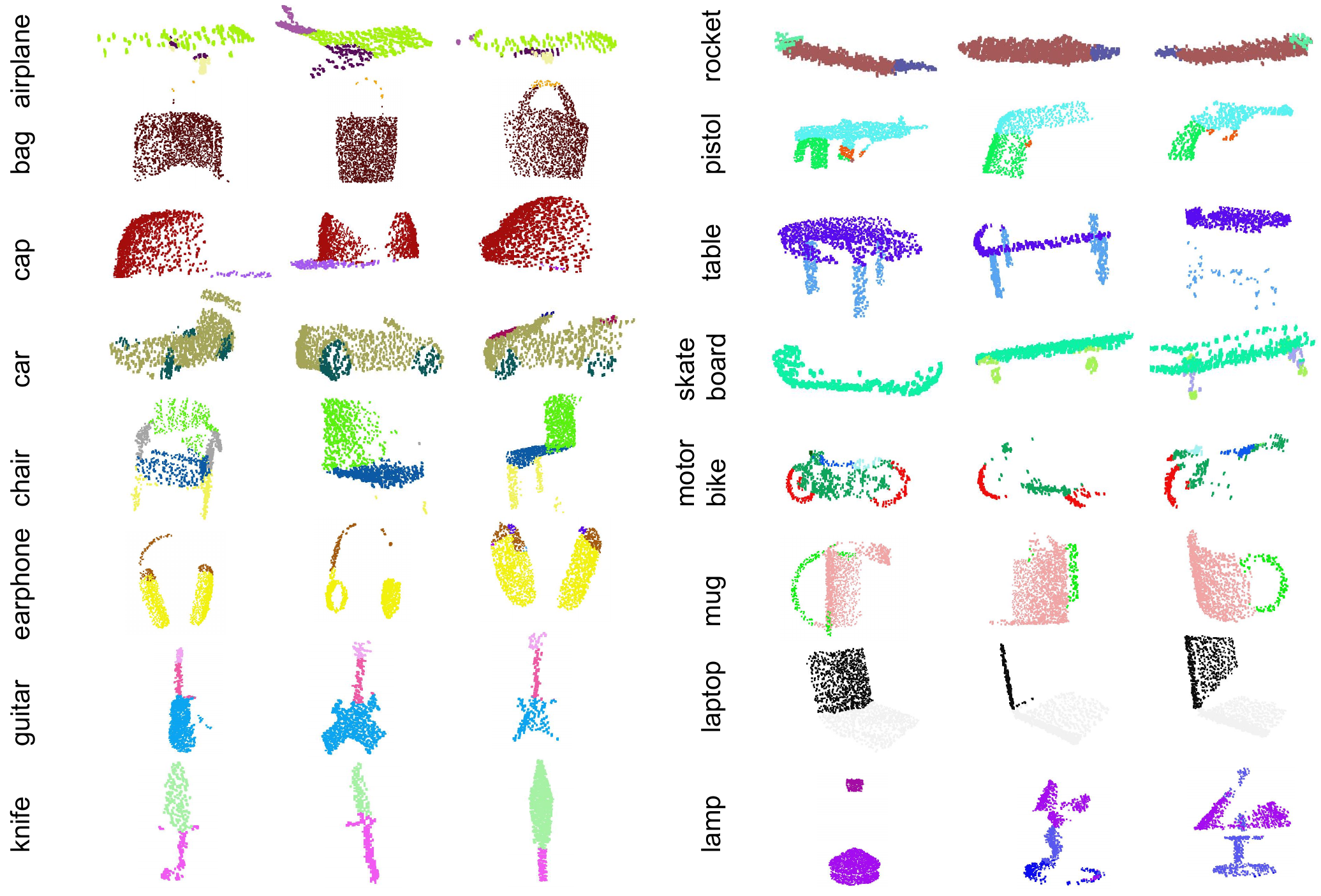}
\caption{\textbf{PointNet segmentation results on simulated Kinect scans.} }
\label{fig:part_seg_partial}
\end{figure*}

\begin{figure*}[t!]
\centering
\includegraphics[width=\linewidth]{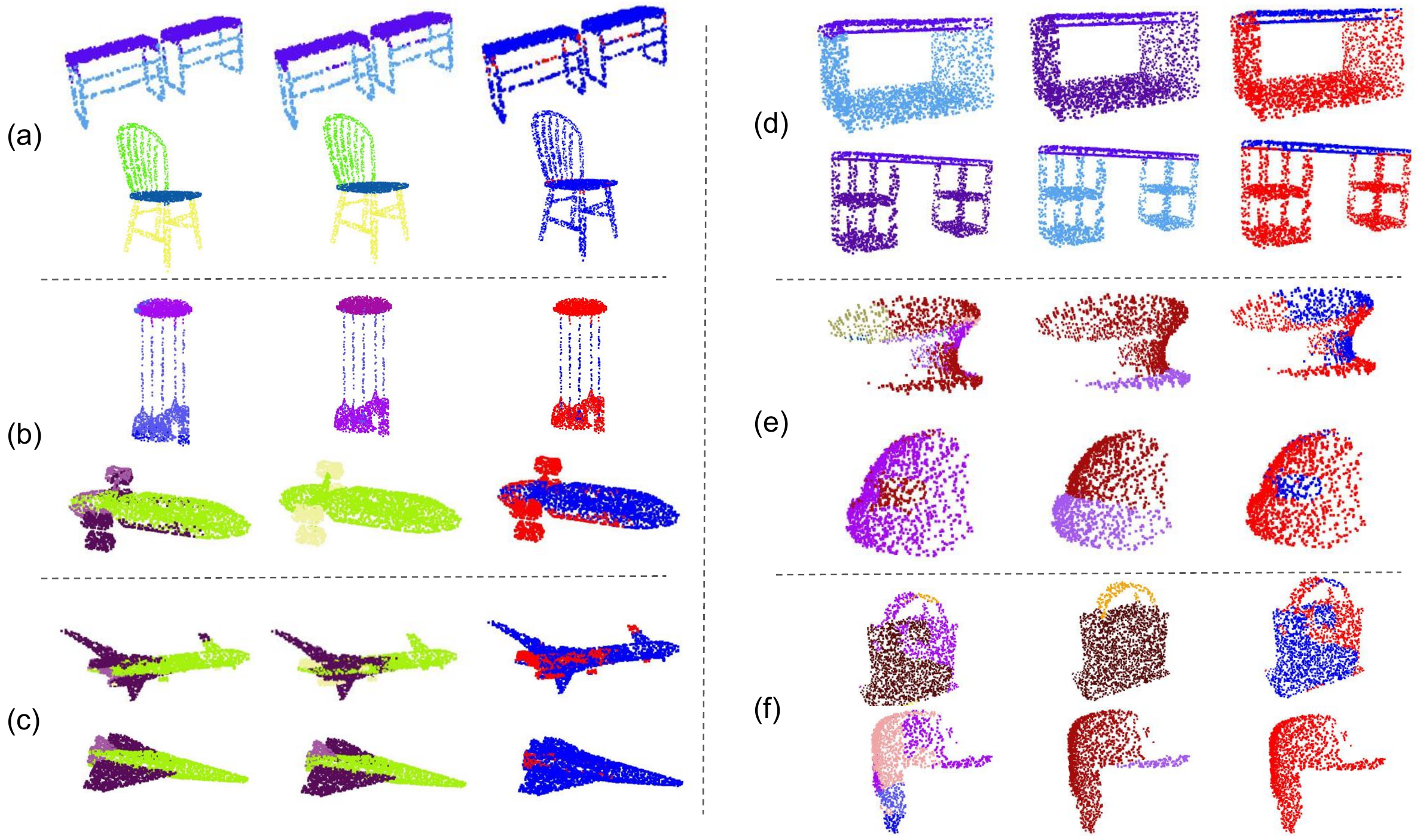}
\caption{\textbf{PointNet segmentation failure cases.} In this figure, we summarize six types of common errors in our segmentation application. The prediction and the ground-truth segmentations are given in the first and second columns, while the difference maps are computed and shown in the third columns. The red dots correspond to the wrongly labeled points in the given point clouds. (a) illustrates the most common failure cases: the points on the boundary are wrongly labeled. In the examples, the label predictions for the points near the intersections between the table/chair legs and the tops are not accurate. However, most segmentation algorithms suffer from this error. (b) shows the errors on exotic shapes. For examples, the chandelier and the airplane shown in the figure are very rare in the data set. (c) shows that small parts can be overwritten by nearby large parts. For example, the jet engines for airplanes (yellow in the figure) are mistakenly classified as body (green) or the plane wing (purple). (d) shows the error caused by the inherent ambiguity of shape parts. For example, the two bottoms of the two tables in the figure are classified as table legs and table bases (category \textit{other} in \cite{Yi16}), while ground-truth segmentation is the opposite. (e) illustrates the error introduced by the incompleteness of the partial scans. For the two caps in the figure, almost half of the point clouds are missing. (f) shows the failure cases when some object categories have too less training data to cover enough variety. There are only 54 bags and 39 caps in the whole dataset for the two categories shown here.}
\label{fig:part_seg_failure}
\end{figure*}

\paragraph{Segmentation Visualization} We present more segmentation results on both complete CAD models and simulated Kinect partial scans. We also visualize failure cases with error analysis. Fig~\ref{fig:part_seg_complete} and Fig~\ref{fig:part_seg_partial} show more segmentation results generated on complete CAD models and their simulated Kinect scans. Fig~\ref{fig:part_seg_failure} illustrates some failure cases. Please read the caption for the error analysis.

\begin{figure*}
    \centering
    \includegraphics[width=\linewidth]{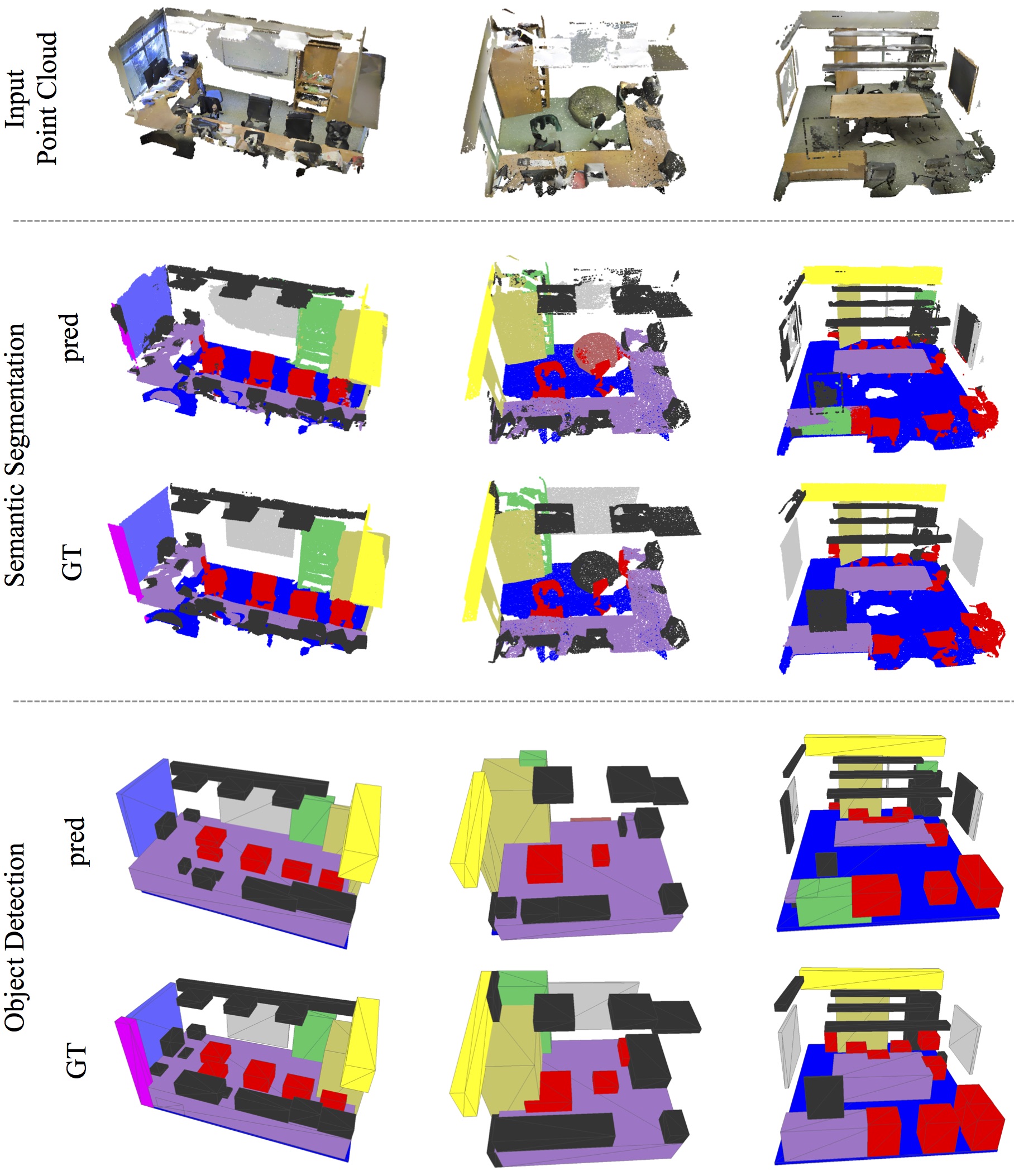}
    \caption{\textbf{Examples of semantic segmentation and object detection.} First row is input point cloud, where walls and ceiling are hided for clarity. Second and third rows are prediction and ground-truth of semantic segmentation on points, where points belonging to different semantic regions are colored differently (chairs in red, tables in purple, sofa in orange, board in gray, bookcase in green, floors in blue, windows in violet, beam in yellow, column in magenta, doors in khaki and clutters in black). The last two rows are object detection with bounding boxes, where predicted boxes are from connected components based on semantic segmentation prediction.}
    \label{fig:semantic_large}
\end{figure*}

\paragraph{Scene Semantic Parsing Visualization}
We give a visualization of semantic parsing in Fig~\ref{fig:semantic_large} where we show input point cloud, prediction and ground truth for both semantic segmentation and object detection for two office rooms and one conference room. The area and the rooms are unseen in the training set.

\paragraph{Point Function Visualization} Our classification PointNet computes $K$ (we take $K=1024$ in this visualization) dimension point features for each point and aggregates all the per-point local features via a max pooling layer into a single $K$-dim vector, which forms the global shape descriptor. 

To gain more insights on what the learnt per-point functions $h$'s detect, we visualize the points $p_i$'s that give high per-point function value $f(p_i)$ in Fig~\ref{fig:functions}. This visualization clearly shows that different point functions learn to detect for points in different regions with various shapes scattered in the whole space.





\end{document}